\documentclass[10pt,twocolumn,letterpaper]{article}

\usepackage{cvpr}              % To produce the CAMERA-READY version
\definecolor{mygray}{gray}{.9}

\usepackage{colortbl}
\usepackage{amssymb}
\usepackage{pifont}
\usepackage{booktabs,multirow}
\usepackage{makecell}
\usepackage{tabulary}
\usepackage{fontawesome5}
\usepackage{bbding}

\newlength\savewidth

\newcommand{\tablestyle}[2]{\setlength{\tabcolsep}{#1}\renewcommand{\arraystretch}{#2}\centering\footnotesize}

\usepackage{amsmath}
\usepackage{capt-of}
\usepackage[dvipsnames]{xcolor}
\usepackage{xspace}
\usepackage{enumitem}
\usepackage{amsmath,amssymb,amsbsy,amsfonts,dsfont,pifont,bm,bbm,mathrsfs,mathtools,nicefrac}
\usepackage{algorithm,algpseudocode,listings}
\usepackage{booktabs,multirow,adjustbox,diagbox,threeparttable,tabularray}

\newcommand{\methodname}{PhysRVG}
\newcommand{\benchname}{PhysRVGBench}
\newcommand{\unifiedname}{MDcycle}
\newcommand{\method}{\texttt{\methodname}\xspace}
\newcommand{\unipost}{\texttt{\unifiedname}\xspace}
\newcommand{\bench}{\texttt{\benchname}\xspace}

\newcommand{\tocite}[1]{{\color{red} [TO CITE]}}

\definecolor{CQColor}{rgb}{0.0,0.0,1.0} % color for Aaron

\definecolor{TSColor}{rgb}{0.5,0.0,0.8} % color for Aaron

\definecolor{CQRColor}{rgb}{1.0,0.0,1.0} % color for Aaron

\definecolor{cvprblue}{rgb}{0.21,0.49,0.74}
\usepackage[pagebackref,breaklinks,colorlinks,allcolors=cvprblue]{hyperref}
\usepackage{wrapfig}
\usepackage[capitalize]{cleveref}  % Should be loaded after 'hyperref', and works perfectly with 'subfigure'.
\crefname{section}{Sec.}{Secs.}
\Crefname{section}{Section}{Sections}
\crefname{table}{Tab.}{Tabs.}
\Crefname{table}{Table}{Tables}
\crefname{figure}{Fig.}{Figs.}
\Crefname{figure}{Figure}{Figures}
\crefname{equation}{Eq.}{Eqs.}
\Crefname{equation}{Equation}{Equations}
\definecolor{baseColor}{rgb}{0.75,0.05,0.1}
\newcommand{\base}[1]{{\color{baseColor}#1}}
\definecolor{checkmarkColor}{rgb}{0.1,0.75,0.1}
\newcommand{\checkc}[1]{{\color{checkmarkColor}#1}}
\definecolor{demphcolor}{RGB}{144,144,144}

\usepackage{tabularx}

\definecolor{cvprblue}{rgb}{0.21,0.49,0.74}
\usepackage[pagebackref,breaklinks,colorlinks,allcolors=cvprblue]{hyperref}

\def\paperID{2971} % *** Enter the Paper ID here
\def\confName{CVPR}
\def\confYear{2026}

\title{PhysRVG: Physics-Aware Unified Reinforcement Learning \\ for Video Generative Models}

\author{Qiyuan Zhang$^{1,2}$\footnotemark[1]\>\,, Biao Gong$^{2}$\footnotemark[2]\>\,, Shuai Tan$^{2}$, Zheng Zhang$^{2}$, Yujun Shen$^{2}$, Xing Zhu$^{2}$,\\ Yuyuan Li$^{1}$, Kelu Yao$^{3}$, Chunhua Shen$^{1}$, Changqing Zou$^{1,3}$\footnotemark[3]\>\,\\[5pt]
{$^1$Zhejiang University\ \ $^2$Ant Group \ \ $^3$Zhejiang Lab} \\
}

\begin{document}
% \maketitle
%%%%%%%%%%%%%%%%%%%%%%%%%%%%%%%%%%%%%%%%%%
\twocolumn[{
\maketitle
\begin{center}
    \vspace{-2pt}
    % \placeholder{17cm}{9.6cm}
    \includegraphics[width=1\linewidth]{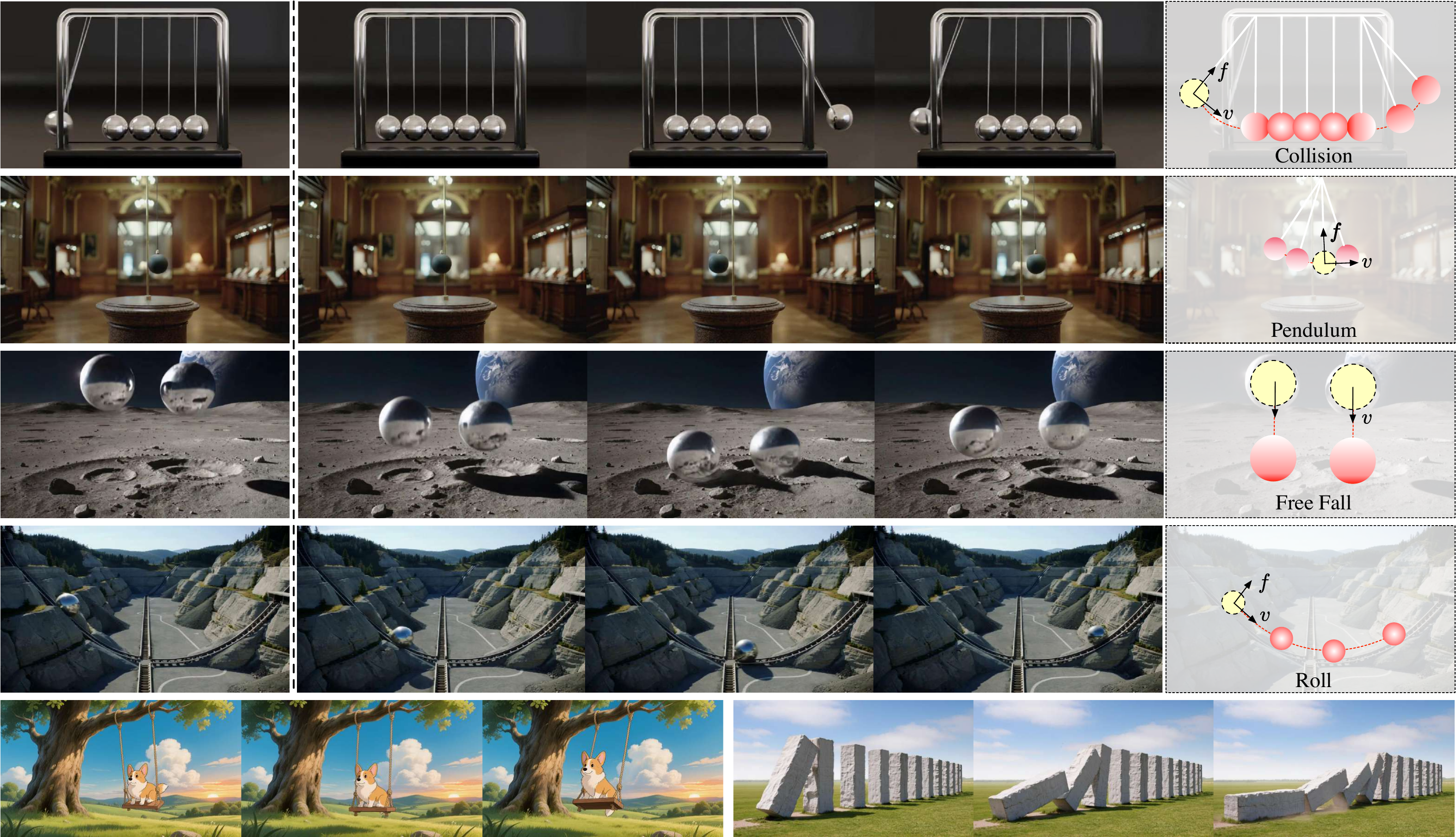}
        \vspace{-6mm}
    \captionof{figure}{\textbf{Samples generated by \method.} Our model produces videos with physically plausible rigid body dynamics. \textit{Rows1–4} display four fundamental types of motion addressed in our work, \textit{row5} validates the model’s generalization in out-of-distribution scenarios.}
    % \vspace{2mm}
    \label{fig:teaser}
\end{center}
}]
%%%%%%%%%%%%%%%%%%%%%%%%%%%%%%%%%%%%%%%%%%

\footnotetext[1]{Work done during internship at Ant Group. \footnotemark[2]Project lead.}
\footnotetext[3]{Corresponding author.}

\begin{abstract}

\vspace{-3mm}
\noindent
% \gb{
Physical principles are fundamental to realistic visual simulation, but remain a significant oversight in transformer-based video generation. 
This gap highlights a critical limitation in rendering rigid body motion, a core tenet of classical mechanics. 
While computer graphics and physics-based simulators can easily model such collisions using Newton formulas, modern pretrain-finetune paradigms discard the concept of object rigidity during pixel-level global denoising. 
Even perfectly correct mathematical constraints are treated as suboptimal solutions (\textit{i.e.}, conditions) during model optimization in post-training, fundamentally limiting the physical realism of generated videos.
Motivated by these considerations, we introduce, for the first time, a physics-aware reinforcement learning paradigm for video generation models that enforces physical collision rules directly in high-dimensional spaces, ensuring the physics knowledge is strictly applied rather than treated as conditions.
Subsequently, we extend this paradigm to a unified framework, termed
% \textbf{Mimicry-Discovery (MD) cycle}
\textbf{Mimicry-Discovery Cycle} (\unipost)
, which allows substantial fine-tuning while fully preserving the model’s ability to leverage physics-grounded feedback.
To validate our approach, we construct new benchmark \bench and perform extensive qualitative and quantitative experiments to thoroughly assess its effectiveness. Our code and ckpt will be released publicly soon. Project page: \href{https://lucaria-academy.github.io/PhysRVG/}{https://lucaria-academy.github.io/PhysRVG/}

% }
\end{abstract}
\vspace{-20pt}

% Physical principles are fundamental to realistic visual simulation, but remain a significant oversight in transformer-based video generation. This gap highlights a critical limitation in rendering rigid body motion, a core tenet of classical mechanics. While computer graphics and physics-based simulators can easily model such collisions using Newton formulas, modern pretrain-finetune paradigms discard the concept of object rigidity during pixel-level global denoising. Even perfectly correct mathematical constraints are treated as suboptimal solutions (i.e., conditions) during model optimization in post-training, fundamentally limiting the physical realism of generated videos. Motivated by these considerations, we introduce, for the first time, a physics-aware reinforcement learning paradigm for video generation models that enforces physical collision rules directly in high-dimensional spaces, ensuring the physics knowledge is strictly applied rather than treated as conditions. Subsequently, we extend this paradigm to a unified framework, termed Mimicry-Discovery Cycle (MDcycle), which allows substantial fine-tuning while fully preserving the model’s ability to leverage physics-grounded feedback. To validate our approach, we construct new benchmark PhysRVGBench and perform extensive qualitative and quantitative experiments to thoroughly assess its effectiveness. Our code and ckpt will be released publicly soon. Project page: \href{https://lucaria-academy.github.io/PhysRVG/}{https://lucaria-academy.github.io/PhysRVG/}

\section{Introduction}
\label{sec:intro}

% \gb{
Physical principles form the foundation of realistic visual simulation, governing how objects move, interact, and respond to forces in the real world~\cite{banerjee2024physicsinformedcomputervisionreview,liu2025generativephysicalaivision,meng2025groundingcreativityphysicsbrief,hu2025simulatingrealworldunified}. 
From rigid body motion to the deformation of soft materials, physical laws establish the continuity and coherence that make dynamic scenes appear physically plausible.
In traditional computer graphics and simulation pipelines~\cite{10.1145/2897826.2927348,xie2023physgaussian}, such principles are explicitly encoded through Newtonian mechanics and numerical solvers~\cite{10.1145/2994258.2994272,10.1016/j.jvcir.2007.01.005}, ensuring that visual outcomes remain physically consistent.

However, recent advances in transformer-based video generation have shifted the focus toward data-driven synthesis, where realism emerges statistically from large-scale training rather than from explicit physical modeling~\cite{ai2025magi1autoregressivevideogeneration,chen2025sanavideoefficientvideogeneration,chen2025skyreelsv2infinitelengthfilmgenerative,ma2025stepvideot2vtechnicalreportpractice,genmo2024mochi}. 
While this paradigm has enabled remarkable progress in semantic understanding and visual fidelity~\cite{lin2025exploringevolutionphysicscognition,meng2025groundingcreativityphysicsbrief}, it remains inherently limited in representing the underlying dynamics of physical real motion. 
The absence of embedded physical constraints leads to inconsistencies such as unstable trajectories~\cite{kong2024hunyuanvideo}, implausible collisions~\cite{ai2025magi1autoregressivevideogeneration}, and a lack of temporal coherence~\cite{bansal2025videophy2challengingactioncentricphysical}, especially when modeling rigid body motion governed by classical mechanics~\cite{liu2024physgen,li2025pisa}.

Despite the importance of physical laws, transformer-based video generators lack explicit structural grounding. Their pretrain-finetune paradigm prioritizes pixel-level reconstruction and perceptual quality, disregarding the constraints imposed by object rigidity.
In addition, current video generative models and training frameworks~\cite{yin2025causvid,huang2025selfforcing,cui2025self} treat physical laws as auxiliary constraints rather than core principles, causing the network to prioritize statistical alignment over physical consistency.
During scaling-up~\cite{wan2025wanopenadvancedlargescale,chen2025sanavideoefficientvideogeneration,kong2024hunyuanvideo}, such alignment is inherently fragile and can be easily compromised, as optimization objectives tend to favor distributional similarity over physical realism.

\begin{figure}[t]
  \centering
    % \placeholder{8cm}{4cm}
    \includegraphics[width=1\linewidth]{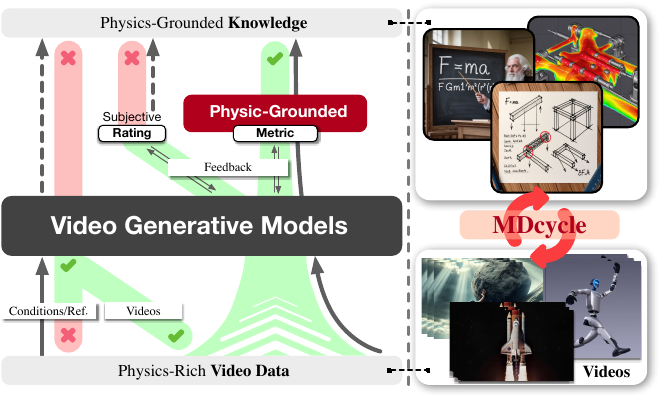}
    \caption{\textbf{The core idea of \method.} DiT-based video generative models reconstruct noisy videos in latent space using Flow Matching loss, which only captures data distributions~(\checkc{\small \checkmark}) but discards essential spatio-temporal physical cues during conditional alignment and feature extraction~(\base{\tiny \XSolidBrush}), thereby hindering stable learning of physical knowledge~(\base{\tiny \XSolidBrush}). While reinforcement learning with subjective ratings can train on physics-rich video data using RL-based feedback~(\checkc{\small \checkmark}), its evaluation remains perceptually biased and fails to provide stable physical supervision~(\base{\tiny \XSolidBrush}). In contrast, our \method leverages the \unipost to fully utilize data for visual refinement~(\checkc{\small \checkmark}) and enforces physical knowledge injection through the Physics-Grounded Metric~(\checkc{\small \checkmark}), enabling stable retention and active discovery of physical principles for truly physics-aware learning and generation~(\checkc{\small \checkmark}).}
    \label{fig:moti}
    \vspace{-4mm}
\end{figure}

% introduce \method. 
Motivated by these insights, as illustrated in Fig.~\ref{fig:moti}, we for the first time introduce \method, a physics-aware reinforcement learning paradigm that enforces physical collision rules directly in high-dimensional spaces for video generative models.
In this work, we focus on \textit{Rigid Body Motion}~\cite{liu2024physgen} as the core representation of physical behavior.
To capture the physical dynamic, we introduce a physics-grounded reward function that integrates motion masks, trajectory offsets, and collision detection into reinforcement learning, enabling models to internalize accurate physical knowledge.
Subsequently, we extend \method into a unified framework through the proposed \textbf{Mimicry-Discovery Cycle} (\unipost). 
It allows the model to undergo substantial parameter adaptation in the early stages, leveraging the \textit{``Mimicry''} phase to capture visual patterns from the data and address unreliable reward signals in standard reinforcement learning. 
A detailed analysis of these reward inconsistencies is provided in Sec.~\ref{sec:md_cycle}. As training progresses, the \textit{``Discovery''} phase enables the model to progressively internalize physical rules, facilitating a smooth transition from data-driven learning to physics-consistent generation. 
The \textit{``Cycle''} mechanism continuously alternates between \textit{``Mimicry''} and \textit{``Discovery''}, allowing the model to dynamically adjust their balance during training and ultimately converge to a stable physics-aware state.

For comprehensive training and evaluation of \method, we construct \bench, a benchmark comprising 700 videos collected from game footage, online sources, and in-house recordings.
These videos cover four types of rigid body motion: collision, pendulum, free fall, and rolling. For each video, we manually annotate the subject’s coordinates in the first frame and leverage SAM2 to generate motion masks across the entire video. 
Based on these masks, we compute two key evaluation metrics, Intersection over Union (IoU)~\cite{rezatofighi2019generalizedintersectionunionmetric} and Trajectory Offset (TO), which provide precise, quantitative measures of physical plausibility in rigid body dynamics.
Experimental results in Fig.~\ref{fig:teaser} and Sec.~\ref{sec:experiment} validate the effectiveness of our approach across a wide range of physical and visual settings.
% }

%%%%%%%%%%%%%%%%%%%%%%%%%%%%%%%%%%%%%%%%%%%%%%%%%
%%%%%%%%%%%%%%%%%%%%%%%%%%%%%%%%%%%%%%%%%%%%%%%%%

\section{Related Work}
\label{sec:related_work}

\subsection{Physics-Aware Video Generation.}
While modern video generative models~\cite{chen2025sanavideoefficientvideogeneration,kong2024hunyuanvideo,nvidia2025cosmosworldfoundationmodel,yang2025cogvideoxtexttovideodiffusionmodels,gong2024check,shi2024motionstone,wei2025dreamrelation,tan2024mimir} can render highly realistic frames, their grasp of physical laws remains insufficient~\cite{huang2024vbench,lin2025exploringevolutionphysicscognition,xiang2025aligningperceptionreasoningmodeling,bordes2025intphys2benchmarkingintuitive}. Methods to enhance physical fidelity in video generation can be categorized by how they inject knowledge~\cite{aira2024motioncraftphysicsbasedzeroshotvideo,tan2024physmotionphysicsgroundeddynamicssingle,zhang2024physdreamer,cao2024neuralmaterialadaptorvisual,lin2025omniphysgs3dconstitutivegaussians,li2024generativeimagedynamics,gong2024uknow}. The first class conditions generation on explicit knowledge. PhysGen~\cite{liu2024physgen} derives motion sequences from rigid body dynamics simulation, GPT4Motion~\cite{lv2024gpt4motionscriptingphysicalmotions} leverages GPT-4o planning and Blender simulation to obtain edge maps and depth maps, NewtonGen~\cite{Yuan_2025_NewtonGen} and PhysAnimator~\cite{xie2025physanimatorphysicsguidedgenerativecartoon} use optical flow as input guidance. These methods depend heavily on the quality and coverage of the simulator. The second approach expands training data with more physics related examples.  Wisa~\cite{wang2025wisa} builds a dataset containing multi-level basic physics knowledge and finetune the model using MoE. Pisa~\cite{li2025pisa} collects free fall videos and applies post-training to teach the model specific physical behaviors. However, such methods inherently confine the model’s capabilities to the motion modes represented in the curated datasets. The third class learns from feedback. PhyT2V~\cite{xue2025phyt2vllmguidediterativeselfrefinement} employs a MLLM to iteratively evaluate generations and refine textual prompts, while PhysMaster~\cite{ji2025physmastermasteringphysicalrepresentation} uses human feedback to rank samples and applies DPO~\cite{rafailov2023dpo} for preference optimization. These pipelines depend on subjective feedback, while ours couples Trajectory Offset (TO) with GRPO~\cite{shao2024grpo} to provide physics-aware rewards.

\subsection{RL for Generative Model.}
Reinforcement learning has achieved strong success in large language models~\cite{ouyang2022traininglanguagemodelsfollow,yang2024qwen2technicalreport,deepseekai2024deepseekv3technicalreport}. RLHF~\cite{gao2023scalingrlhf} uses human ranking and preferences as feedback and optimizes the model with Proximal Policy Optimization(PPO)~\cite{pposchulman2017proximalpolicyoptimizationalgorithms}. DeepSeek -R1~\cite{deepseekai2025deepseekr1incentivizingreasoningcapability} uses verifiable outcomes as feedback and applies Group Relative Policy Optimization(GRPO)~\cite{shao2024grpo} to greatly improve reasoning. Compared to PPO, GRPO is more efficient because it does not require training a separate value model. Motivated by these successes, DDPO~\cite{black2023trainingddpo} brings reinforcement learning into diffusion-based generative models and formulates the denoising process as a Markov Decision Process(MDP). Recently, Flow-GRPO~\cite{liu2025flowgrpo} and DanceGRPO~\cite{xue2025dancegrpo} push GRPO into flow matching models by converting ODE sampling into equivalent SDE sampling and adding tunable exploration noise. For faster training, MixGRPO~\cite{li2025mixgrpo} proposes a hybrid ODE-SDE strategy. The model reaches competitive results with very few optimization steps and a short training time. TempFlowGRPO~\cite{he2025tempflow} and G$^2$RPO~\cite{zhou2025text} leverage the intrinsic temporal structure of flow-based models and address uneven credit assignment under sparse rewards.

%%%%%%%%%%%%%%%%%%%%%%%%%%%%%%%%%%%%%%%%%%%%%%%%%
\begin{figure*}[t]
  \centering
    % \placeholder{17cm}{8cm}
    \includegraphics[width=1\linewidth]{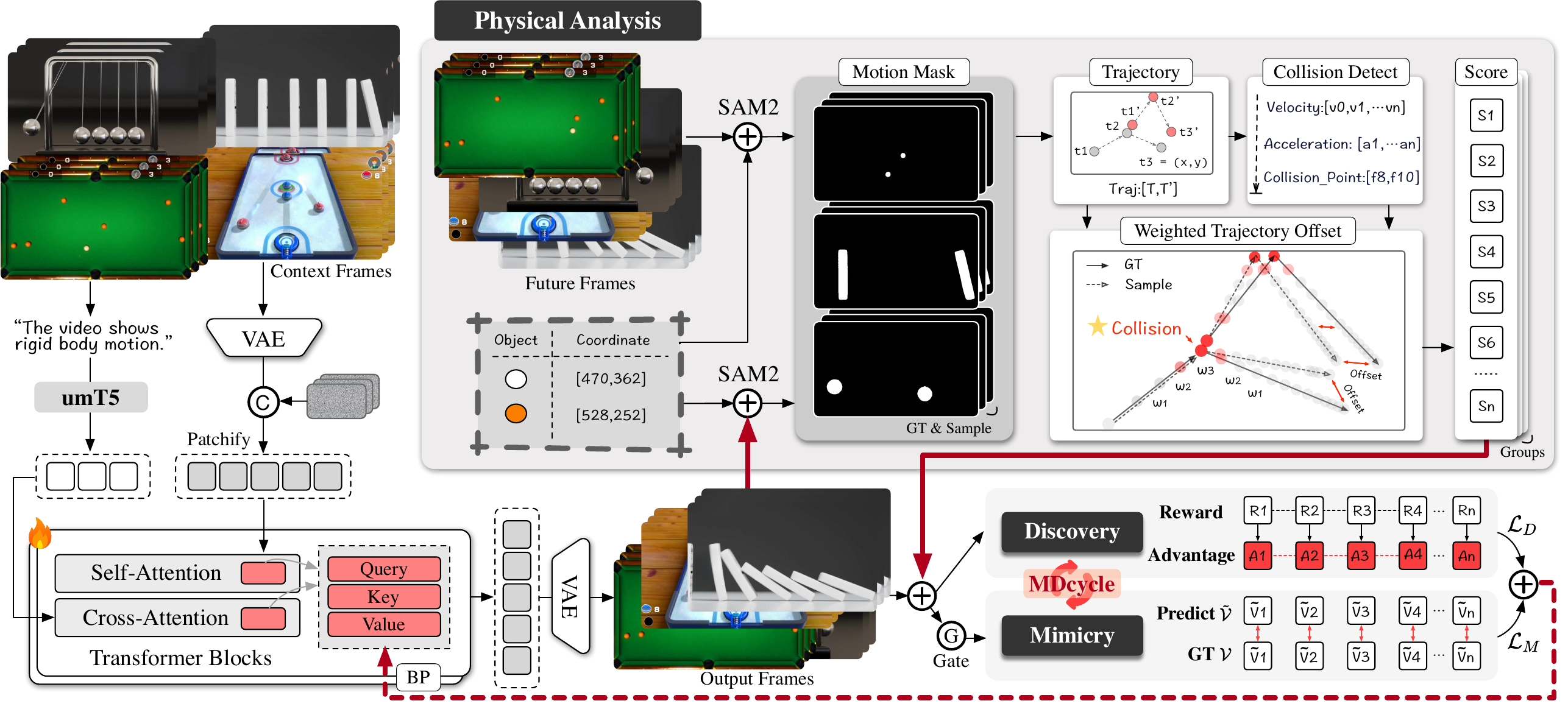}
    \caption{\textbf{The framework of \method.} Given a text prompt and context frames, the model generates future video frames. For both the groundtruth and sampled frames, we derive motion masks $M$ by prompting SAM2~\cite{ravi2024sam2} with object coordinates $p_1$ from the first frame, which are manually annotated during data preprocessing. We then compute object trajectories $P$ and perform collision detection. The trajectory offset $O$ between the sampled and groundtruth trajectories is calculated and reweighted by the collision signal $w_t$ to yield a weighted trajectory offset $O_c$, which serves as the per-sample score. All transformer blocks are trained with full parameters.
    }
    \label{fig:framework}
    \vspace{-2mm}
\end{figure*}
%%%%%%%%%%%%%%%%%%%%%%%%%%%%%%%%%%%%%%%%%%%%%%%%%

\section{Methodology}
\label{sec:method}

In this section, we first present the necessary background on reinforcement learning and flow matching in~\cref{sec:preliminary}. In~\cref{sec:task_definition}, we define our task and articulate its relationship with physical knowledge. Subsequently,~\cref{sec:reward_modeling} details the design of our reward function. We then present \method, a unified post-training method in~\cref{sec:md_cycle}. Finally,~\cref{sec:train_pipeline} outlines the overall training pipeline. The overall framework of \method is illustrated in~\cref{fig:framework}.

\subsection{Preliminary}
\label{sec:preliminary}

\noindent \textbf{Flow Matching.} Flow Matching~\cite{lipman2023flowmatchinggenerativemodeling,liu2022rectifyflow} treats generation as a denoising process from noise to data. Assume $x_0$ is sampled from real data and $x_1$ is sampled from Gaussian noise distribution. The intermediate sample $x_t$ is a linear interpolation of $x_0$ and $x_1$, defined as: $x_t = (1 - t)x_0 + t x_1$. A generative model is trained to predict the velocity field $v = x_1 - x_0$. The optimization objective for this process can be formulated as:
\begin{equation}
L(\theta) = \mathbb{E}_{t, x_0 \sim X_0, x_1 \sim X_1 }[\|v - v_\theta (x_t, t) \|^2].
\end{equation}
\noindent \textbf{Reinforcement Learning.} According to Flow-GRPO~\cite{liu2025flowgrpo} and DanceGRPO~\cite{xue2025dancegrpo}, the ODE governing the Flow Matching denoising trajectory can be convert to SDE while preserving the same marginal distributions, the SDE denoising process can be formulated as:

\begin{equation}
  \mathrm{d} x_{t}=\left[v_{t}+\frac{\sigma_{t}^{2}}{2 t}\left(x_{t}+(1-t) v_{t}\right)\right] \mathrm{d} t+\sigma_{t} \mathrm{~d} w
  \label{eq:sde}
\end{equation}
where $v_t$ is the predicted velocity, $\sigma_t$ is a hyperparameter controlling the strength of stochastic noise (i.e., the intensity of exploration in RL), and $w$ represents standard Brownian motion.

GRPO estimates the advantage value for each sample by comparing it against other samples within the same group. Given a group of $G$ samples $\{x_0^i\}_{i=1}^G$ generated from the same condition $c$, the advantage corresponding to the $i$-th sample is formulated as: 
\begin{equation}
    {A}_t^i=\frac{r(x_0^i, c)-\text{mean}(\{r(x_0^j, c)\}_{j=1}^G)}{\text{std}(\{r(x_0^j, c)\}_{j=1}^G)}
\label{eq:advantage}
\end{equation}

The GRPO algorithm then optimizes the policy model by maximizing the following objective:
\begin{align}
J(\theta) = \frac{1}{G} \sum_{i=1}^{G} \frac{1}{T} \sum_{t=0}^{T-1}
\Big(
\min\!\big(r_t^{i}(\theta)\, \hat{A}_t^{i},\; \notag\\
\text{clip}(r_t^{i}(\theta),\, 1 - \epsilon,\, 1 + \epsilon)\, \hat{A}_t^{i} -D_{KL}(\pi_\theta \,\|\, \pi_{\text{ref}}) \big)
\Big),
\label{eq:rl}
\end{align}

where $r_t^{i}(\theta) = \frac{ p_{\theta}(x_{t-1}^i|x_t^i, \bm{c})}{p_{\theta_{old}}(x_{t-1}^i|x_t^i, \bm{c})}$ serves as an importance sampling ratio, correcting for the bias of evaluating the current policy using data generated by an older one. The $D_{KL}$ term acts as a regularizer to promote training stability.

\subsection{Physics-Grounded Task Definition}
\label{sec:task_definition}
Physical laws provide a natural foundation for defining verifiable tasks in video generation. Among them, rigid-body motion serves as a fundamental and well-defined phenomenon that enables quantitative evaluation. It exhibits two intrinsic properties:
(1) \textit{Observability}: the motion of a rigid body can be precisely represented through coordinate transformations, and its position is directly measurable from video frames;
(2) \textit{Determinism}: given the initial position and velocity, the subsequent trajectory of a rigid body is uniquely determined by Newtonian mechanics. 
These properties make rigid-body motion suitable for verifying the physical plausibility of generated videos. In this work, we consider four representative types of motion, including collision, pendulum, free fall, and rolling. 

Specifically, our task can be formulated as simulating the rigid body dynamics by generating future frames based on the observed initial frames. Given the initial $T_\mathrm{obs}$ frames $\{I_1, I_2, \dots, I_{T_\mathrm{obs}}\}$ and text prompt $c$, we aim to generate the subsequent $T_\mathrm{pred}$ frames $\{I_{T_\mathrm{obs}+1}, \dots, I_{T_\mathrm{obs}+T_\mathrm{pred}}\}$. This can be expressed as:
\begin{equation}
    \label{eq:task_definition}
    \hat{I}_{1:T_\mathrm{obs}+T_\mathrm{pred}}
    = \mathcal{F}_{\theta}\left(I_{1:T_\mathrm{obs}},c\right)
\end{equation}
Building on this definition, we further formulate a physics-grounded reward function that quantitatively measures how well the generated videos aligns with physical laws. The details of the reward modeling are described in~\cref{sec:reward_modeling}.

\subsection{Reward Modeling}
\label{sec:reward_modeling}
To enable physics-aware reinforcement learning, we design a new reward function guided by the principles established in the Physics-Grounded Task Definition.
In reinforcement learning, the key challenge lies in designing reliable feedback for generated outputs, which directly determines the model’s stability and learning direction.
Many previous works~\cite{ji2025physmastermasteringphysicalrepresentation,xue2025phyt2vllmguidediterativeselfrefinement} rely on MLLM or human evaluations as the feedback signals, which are inherently subjective and coarse-grained. 
Such feedback cannot precisely verify the physical plausibility of generated videos and also fails to capture the relative quality among samples.
Therefore, in this paper, we propose a new Physics-Grounded Metric for reward modeling, which consists of two core components: the \textit{Trajectory Offset} and the \textit{Collision Detection}.

\noindent \textbf{Trajectory Offset.}
Since we establish a Physics-grounded Task Definition, where the motion of a rigid body is represented by the variation of its center coordinates, we then denote the trajectory of the object by 
$p_{1:T} = \{p_1, p_2, \dots, p_T\},$
where $p_t \in \mathbb{R}^2$ indicates the center position of the object at frame $t$.
Because rigid-body motion does not involve deformation, the coordinate sequence $p_{1:T}$ can precisely reflect both positional and velocity changes over time.
During data preparation, we manually annotate the object’s center coordinate $p_1$ in the first frame.
In collision scenarios, two objects are annotated, corresponding to the active and passive bodies, while in other scenarios only one object is annotated. 
As shown in~\cref{fig:framework}, given the annotated initial coordinate $p_1$, both the groundtruth and generated videos are processed by SAM2~\cite{ravi2024sam2} to extract motion mask sequences $M_{1:T}^{\text{gt}}$ and $M_{1:T}^{\text{sample}}$. 
We then compute the mean pixel of each mask as the object center:
\begin{equation}
p_{1:T}^{\text{gt}} = \mathrm{Center}(M_{1:T}^{\text{gt}}),  
p_{1:T}^{\text{sample}} = \mathrm{Center}(M_{1:T}^{\text{sample}})
\end{equation}
Lastly, we quantify the Trajectory Offset(TO) between the generated and ground-truth trajectories by computing the frame-wise deviation:
\begin{equation}
    O = \frac{1}{N T} 
        \sum_{s=1}^{N} \sum_{t=T_{obs}}^{T} 
        \left\| 
            p_{t,s}^{\text{gt}} 
            - 
            p_{t,s}^{\text{sample}} 
        \right\|_2
    \label{eq:offset}
\end{equation}
where $N$ denotes the number of annotated rigid bodies in the scene and $p_{t,s}^{\text{gt}}$ and $p_{t,s}^{\text{sample}}$ 
represent the 2D coordinates of the $s$-th object at frame $t$
in the ground-truth and generated videos, respectively.

\noindent \textbf{Collision Detection.} 
Optimizing the model only with the offset in Eq.~\ref{eq:offset} causes reward hacking, where it favors simple linear motions and avoids complex collisions~(\cref{sec:ablation_collision}). To address this, we upweight losses near collision frames, guiding the model to focus on critical physical interactions.
Specifically, given the object trajectory $p_{1:T} = \{p_1, p_2, \dots, p_T\}$, we compute the velocity sequence $v_{2:T}$, where each term is defined as $v_n = \frac{p_n - p_{n-1}}{\Delta t}$, Then, the acceleration sequence $a_{3:T}$ is derived by $a_n = \frac{v_n - v_{n-1}}{\Delta t}.$ According to Newton's second law $F=ma$, the occurrence of a collision can be detected by sudden changes in acceleration. The detection method is provided in our \textit{Appendix}. We identify the set of collision timestamps $\mathcal{C}$ and their adjacent timestamps 
$\mathcal{C}_{\mathrm{adj}} = \{ t \pm 1 \mid t \in \mathcal{C} \} \setminus \mathcal{C}$. 
To emphasize these critical moments, a temporal weight $w_t$ is assigned to each frame $t$ as follows:
\begin{equation}
w_t =
\begin{cases}
    w^{col}, & \text{if } t \in \mathcal{C} \\
    w^{adj},  & \text{if } t \in \mathcal{C}_\text{adj} \\
    w, & \text{otherwise}
\end{cases}
\label{eq:temporal_weights}
\end{equation}
$w^{col}$,$w^{adj}$and $w$ are hyperparameters, we provide an ablation study of their effects in Sec.~\ref{sec:ablation_collision} and Fig.~\ref{fig:ablation_mdcycle}. Finally, the weighted trajectory offset is expressed as:
\begin{equation}
    O_\text{c} = 
    \frac{1}{N T} 
    \sum_{s=1}^{N} \sum_{t=T_\text{obs}}^{T} 
    w_t
    \left\| 
        p_{t,s}^{\text{gt}} 
        - 
        p_{t,s}^{\text{sample}} 
    \right\|_2,
    \label{eq:weighted_offset}
\end{equation}
The offset $O_\text{c}$ measures trajectory discrepancies between generated and groundtruth motions, where smaller values indicate higher accuracy and stronger physical consistency. 
During reinforcement learning, the reward is defined as its negative, $R = -O_\text{c}$, so that higher rewards correspond to smaller trajectory errors, guiding the policy toward physically accurate and temporally coherent motion generation.

%%%%%%%%%%%%%%%%%%%%%%%%%%%%%%%%%%%%%%%%%%%%%%%%%
\begin{figure}[t]
  \centering
    % \placeholder{8cm}{4cm}
    \includegraphics[width=1\linewidth]{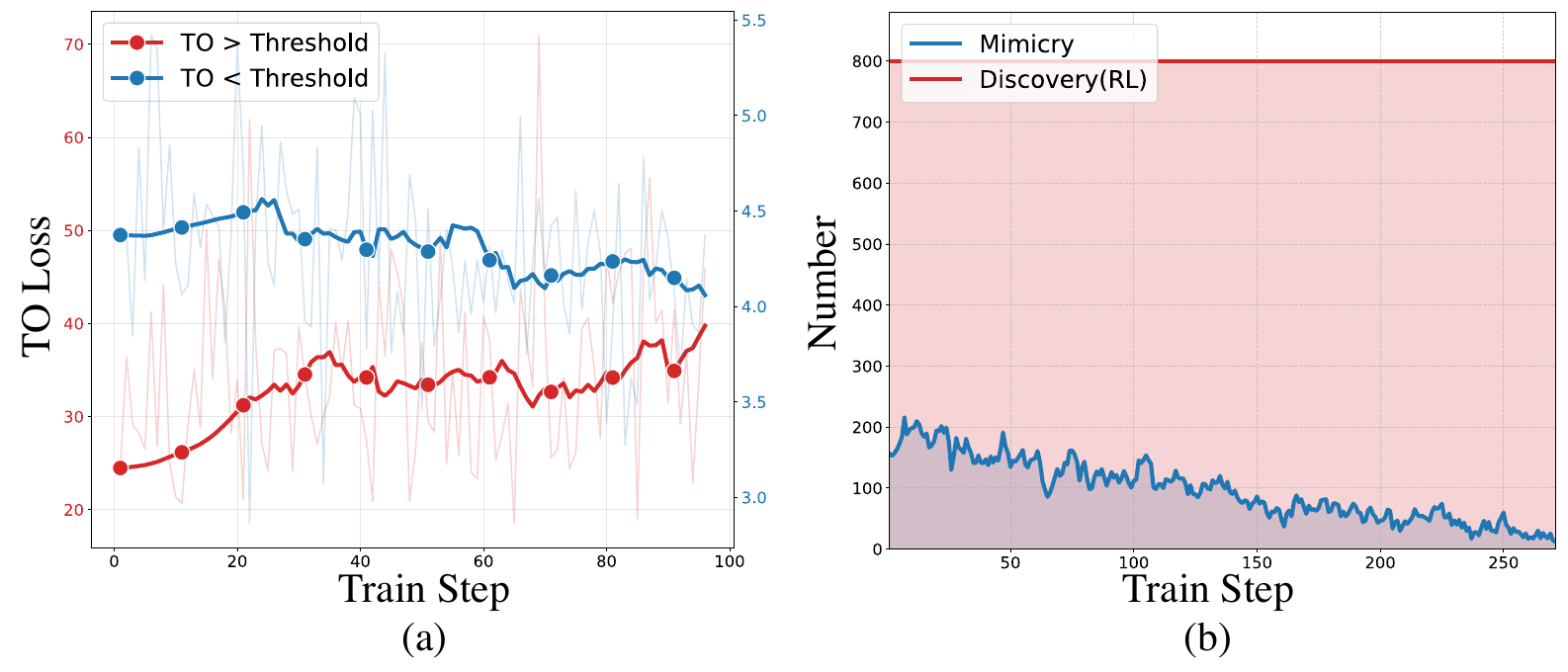}
    \vspace{-5mm}
    \caption{(a) \textbf{RL Training loss} for samples of varying quality. (b) \textbf{Number of samples} assigned to the Mimicry and Discovery branches throughout \unipost.}
    \label{fig:method_motivation}
    \vspace{-2mm}
    % \vspace{-1mm}
\end{figure}
%%%%%%%%%%%%%%%%%%%%%%%%%%%%%%%%%%%%%%%%%%%%%%%%%

\begin{table*}[!t]
\begin{center}
\setlength{\tabcolsep}{2.5mm}
\renewcommand\arraystretch{0.9} 
% \small
\resizebox{\linewidth}{!}{
\begin{tabular}{l| cccccccc|cc|cc}
\toprule
Model & \makecell{Subj.\\Cons.}$\uparrow$ & \makecell{Back.\\Cons.}$\uparrow$ & \makecell{Image\\Qual.}$\uparrow$  & \makecell{Moti.\\Smoo.}$\uparrow$ & \makecell{Dyna.\\Degr.}$\uparrow$ & \makecell{Aest.\\Qual.}$\uparrow$ & \makecell{Temp.\\Flic.}$\uparrow$ & \makecell{Total\\Score}$\uparrow$ & SA$\uparrow$ & PC$\uparrow$ & \makecell{IoU}$\uparrow$ & \makecell{TO}$\downarrow$ \\
\midrule\\[-4.4mm]
\rowcolor{mygray}\multicolumn{13}{l}{\textit{I2V models}}  \\
Wan2.2 5B     & 86.64 & 90.35 & 59.06 & 97.87 & \underline{56.00} & 37.02 & 97.11 & 74.86 & 0.57 & 0.21 & 0.15 & 162.78 \\
Wan2.2 14B    & 94.75 & 95.94 & 60.66 & 99.21 & 50.00 & 38.61 & 98.64 & 76.83 & 0.64 & 0.34 & 0.12 & 162.40 \\
Kling2.5      & 95.32 & 95.56 & \underline{64.85} & 99.47 & \textbf{57.14} & 40.29 & 98.76 & \underline{78.77} & \underline{0.70} & \underline{0.41} & 0.23 & \underline{103.22} \\
Cogvideox     & 92.28 & 95.87 & 52.49 & 99.10 & 46.00 & 38.89 & 98.66 & 74.76 & 0.56 & 0.19 & 0.12 & 158.01 \\
HunyuanVideo  & 95.99 & 96.75 & 59.80 & 99.47 & 46.00 & \underline{41.28} & 99.06 & 76.97 & 0.60 & 0.32 & 0.10 & 181.62 \\
\rowcolor{mygray}\multicolumn{13}{l}{\textit{V2V models}}  \\
Magi-1        & \textbf{97.03} & \underline{97.40} & 60.66 & \underline{99.57} & 48.28 & 37.67 & \underline{99.55} & 77.16 & 0.67 & 0.38 & \underline{0.27} & 113.42 \\

\midrule 
\method       & \underline{96.97} & \textbf{97.74} & \textbf{64.96} & \textbf{99.63} & 52.00 & \textbf{41.36} & \textbf{99.58} & \textbf{78.89} & \textbf{0.76} & \textbf{0.44} & \textbf{0.64} & \textbf{15.03} \\

\bottomrule
\end{tabular}
}
\vspace{-2mm}
\caption{\textbf{Quantitative comparisons} with existing methods on Vbench, VideoPhy-2 (SA, PC) and \bench (IoU, TO). The best results are shown in \textbf{bold}, and the second best are \underline{underlined}.}
\vspace{-4mm}
\label{tab:quantity_exp}
\end{center}
\end{table*}

\subsection{Mimicry-Discovery Cycle}
\label{sec:md_cycle}

Despite being capable of reinforcement learning, the model encounters two practical challenges: (1) difficulty in convergence with small batches and (2) instability in early training stages even with large ones~(\cref{fig:ablation_mdcycle}).
As shown in~\cref{fig:method_motivation}, when trained purely with reinforcement learning, the model tends to perform worse on challenging samples even as it continues to improve on easier ones. This phenomenon arises because the model fails to generate high-quality outputs within a limited number of exploratory attempts, leaving it insufficiently prepared for effective exploration in reinforcement learning.
To address this issue, we apply the Flow Matching loss to these challenging cases, providing fine-grained pixel-level supervision that stabilizes early training and improves convergence.

Specifically, given a textual condition $c$ and a context video $I_{1:T_\mathrm{obs}}$, the model generates a group of video samples $\{x_o^i\}_{i=1}^G$. Each sample is evaluated by the evaluate function $O_c$ described above, and we compute per-sample advantage value $\{A^i\}_{i=1}^G$ using~\cref{eq:advantage}.  Additionally, we compute the group-average trajectory offset $\bar{O}_c = \frac{1}{G} \sum_{i=1}^G O_c(x_0^i, c)$ to assess the performance on this group.
We introduce a hyperparameter $\mathrm{Threshold}$ to control the strategy switch:  when $\bar{O}_c > \mathrm{Threshold}$, it indicates that the model performs poorly on this case, and we add an additional Flow Matching loss term $L_M$ to guide optimization. Otherwise, the training relies primarily on the RL objective $L_D = -J(\theta)$~\cref{eq:rl}. Formally, the unified optimization objective is defined as:
\vspace{-3mm}
\begin{align}
    L &= L_D + \alpha L_M, \label{eq:md_cycle_loss} \\
    \text{where} \quad \alpha &=
    \begin{cases}
        1, & \text{if } \bar{O}_c > \mathrm{Threshold} \\
        0, & \text{otherwise.}
    \end{cases} \notag
\end{align}
\unipost essentially operates under a reinforcement learning framework, which is more complex than standard fully finetune or PEFT such as LoRA. Due to space limitations, we provide only a brief description here and include the full \textit{pseudocode} in the \textit{Appendix}.

\subsection{Training Pipeline}
\label{sec:train_pipeline}
To implement \unipost in practice, we follow a two-stage training pipeline that gradually transitions the model from conventional image-to-video generation to physics-aware video-to-video generation.  
We start from a pretrained diffusion transformer with strong visual priors and adapt it to V2V in the first stage. Specifically, we replace the original image condition with the first $T_{\mathrm{obs}}=5$ frames of the input video and perform full-parameter fine-tuning on our training set using the Flow Matching loss, enabling basic temporal generation capability.  
In the second stage, we focus on improving physical consistency through training under the \unipost framework, using a parameter-efficient weight setup inspired by LoRA and reinforcement feedback guided by the Physics-Grounded Metric.
This staged pipeline ensures stable optimization while introducing physics-aware objectives. More details and ablation studies of the training schedule are provided in \textit{Appendix},~\cref{subtab:training_schedule}, and~\cref{sec:ablation_study}.

\section{Experiment}
\label{sec:experiment}

\subsection{Experimental Settings}
\label{sec:experimental_settings}

\label{sec:dataset}
\noindent \textbf{Dataset.}
Our training data consist of both open-source and proprietary video collections, totaling approximately 10M samples. 
The open-source portion includes Panda-70M, InternVid, and WebVid-10M, providing diverse motion scenes and visual contexts.
To supplement these, we additionally collect proprietary video data, including real-world competition footage, synthetic videos recorded in video games, and real experiments captured using our own recording equipment. For high-quality rigid-body motion data, which are relatively scarce, we manually curate and annotate about 700 video samples covering various motion types such as collision, rolling, pendulum, and free fall. Among them, around 50 videos are separated as the benchmark evaluation set, which is completely excluded from training. Representative samples and annotation examples are provided in the \textit{Appendix}.

\label{sec:evaluation_metrics}
\noindent \textbf{Evaluation Metrics.}
We adopt VBench~\cite{huang2024vbench} to evaluate visual fidelity and VideoPhy-2~\cite{bansal2025videophy2challengingactioncentricphysical} to evaluate physical realism in generated videos.
To further evaluate physical realism in the rigid body motion, we employ \bench, which measures the discrepancy between generated and ground-truth motions using Intersection-over-Union (IoU) and Trajectory Offset (TO) to capture spatial overlap and trajectory deviation.
Details of our newly proposed IoU and TO are provided in \textit{Appendix}.

\subsection{Experimental Results}
\label{sec:experimental_results}

\noindent \textbf{Quantitative Comparisons.} 
Table~\ref{tab:quantity_exp} shows that \method delivers high visual quality on VBench and achieves clear gains on VideoPhy-2 and \bench.
It consistently outperforms existing methods in both IoU and TO, highlighting two key observations. \textit{First}, V2V models generally achieve better physical realism than I2V models, since video inputs provide more reliable motion cues than text descriptions. \textit{Second}, \method further improves over the best V2V baseline, confirming its superior capability in simulating rigid body motions.
\label{sec:quantitative_results}

\begin{figure*}[t]
  \centering
    % \placeholder{17cm}{8cm}
    \includegraphics[width=1\linewidth]{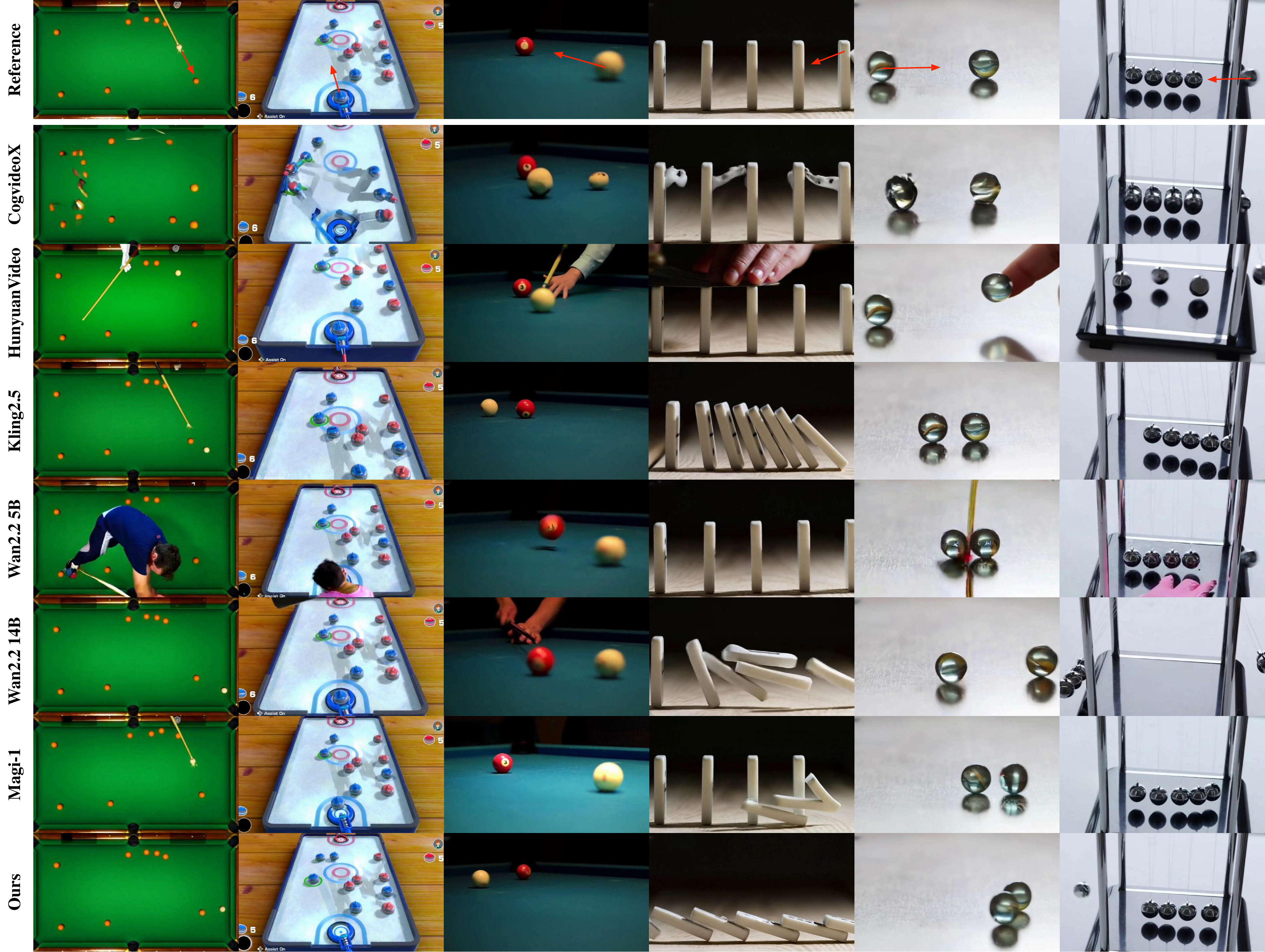}
    \vspace{-6mm}
    \caption{\textbf{Qualitative comparisons} with existing methods. Each sample in the figure corresponds to the final frame of a generated video. We include the original videos for all cases in the \textit{Supplementary Materials}.}
    \label{fig:quality_comparison}
    \vspace{-4mm}
\end{figure*}

\noindent \textbf{Qualitative Comparisons.} 
\label{sec:quality_results}
As shown in~\cref{fig:quality_comparison}, competing methods fail to reproduce stable rigid body motions. Even in simple linear rolling, objects often follow incorrect trajectories or remain static (column-3,5), and in collision scenes the outputs exhibit tearing, overlap, and unnatural merging (column-1,2,4,6). We observe characteristic failure cases: an unexpected human appears in row-5-column-1, domino pieces overlap and cross in row-7-column-4, and the glass ball changes color and becomes distorted in row-2-column-4. These errors are consistent with a human-centric dataset bias that favors generating people over enforcing physical plausibility. We include the original videos for all cases in the \textit{Supplementary Materials}, and all results are produced with a single random seed without manual selection. In contrast, \method preserves physical integrity and motion coherence across these settings, accurately capturing rolling and collision behaviors and reflecting a sound grasp of rigid body physics.

%%%%%%%%%%%%%%%%%%%%%%%%%%%%%%%%%%%%%%%%%%%%%%%%%
%%%%%%%%%%%%%%%%%%%%%%%%%%%%%%%%%%%%%%%%%%%%%%%%%

%%%%%%%%%%%%%%%%%%%%%%%%%%%%%%%%%%%%%%%%%%%%%%%%%

\begin{table}[t]
\centering
\begin{minipage}{1\linewidth}

\subfloat[
]{
\begin{minipage}{0.5\linewidth}
\tablestyle{4pt}{1.1}
\begin{tabular}{ccc}
\centering
Method & IoU$\uparrow$ & TO$\downarrow$ \\
\hline
Baseline & 0.15 & 162.78 \\
\hline
Baseline+LoRA & 0.41 & 48.60 \\
Baseline+FT    & 0.38 & 46.27 \\
Baseline+FT+RL & 0.61 & 17.25 \\
Baseline+FT+MD & \cellcolor{mygray}\textbf{0.64} & \cellcolor{mygray}\textbf{15.03} \\
\end{tabular}
\label{subtab:training_strategy}
\end{minipage}
}
% ################################################
% ################################################
\subfloat[
]{
\begin{minipage}{0.5\linewidth}
\tablestyle{2.5pt}{1.1}
\begin{tabular}{ccc}
\centering
$(w,w^{adj},w^{col})$ & IoU$\uparrow$ & TO$\downarrow$ \\
\hline
(1,1,1) & 0.64 & 15.41 \\
(1,2,3) & \cellcolor{mygray}\textbf{0.64} & \cellcolor{mygray}\textbf{15.03} \\
(1,2,4) & 0.64 & 15.08 \\
(1,2,5) & 0.63 & 16.17 \\
\end{tabular}
\label{subtab:collision_weight}
\end{minipage}
}
\vspace{1em}
% ################################################
% ################################################
% ################################################
\end{minipage}
\vspace{-7mm}
\caption{(a) \textbf{Ablation Study of Training Strategies}. FT denotes full-parameter fine-tuning, and MD denotes \unipost. (b) \textbf{Hyperparameter Analysis of the Collision Weight}. The best configuration is marked in \colorbox{mygray}{gray}. 
}
\label{tab:ablation}
\vspace{-6mm}
\end{table}
%##################################################################################################
%%%%%%%%%%%%%%%%%%%%%%%%%%%%%%%%%%%%%%%%%%%%%%%%%

\subsection{Ablation Study}
\label{sec:ablation_study}

%%%%%%%%%%%%%%%%%%%%%%%%%%%%%%%%%%%%%%%%%%%%%%%%%
\begin{figure}[t]
  \centering
  \vspace{-2mm}
    % \placeholder{8cm}{4cm}
    \includegraphics[width=1\linewidth]{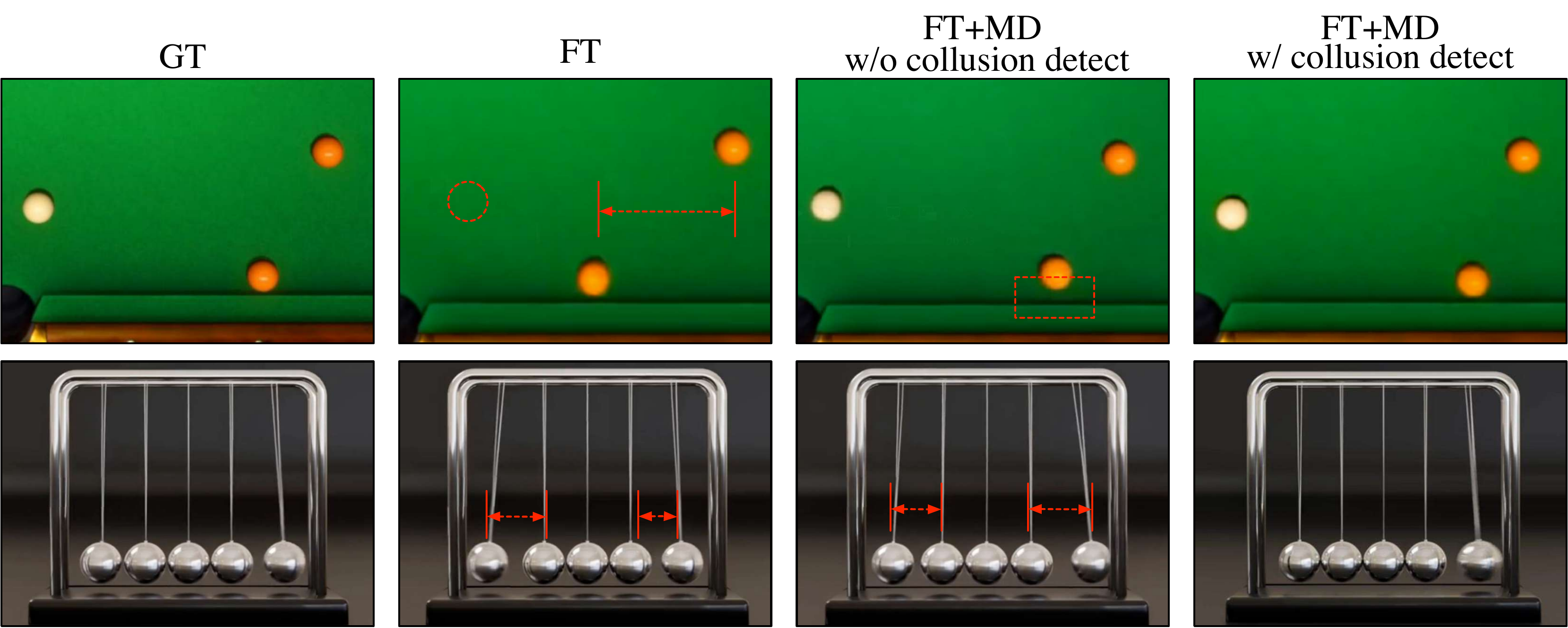}
    \vspace{-7mm}
    \caption{\textbf{Ablation Study of the Collision Detection.}}
    \label{fig:ablation_collision}
    \vspace{-5mm}
\end{figure}
%%%%%%%%%%%%%%%%%%%%%%%%%%%%%%%%%%%%%%%%%%%%%%%%%

%%%%%%%%%%%%%%%%%%%%%%%%%%%%%%%%%%%%%%%%%%%%
\begin{figure*}[!t]
  \centering
    % \placeholder{8cm}{4cm}
    \includegraphics[width=1\linewidth]{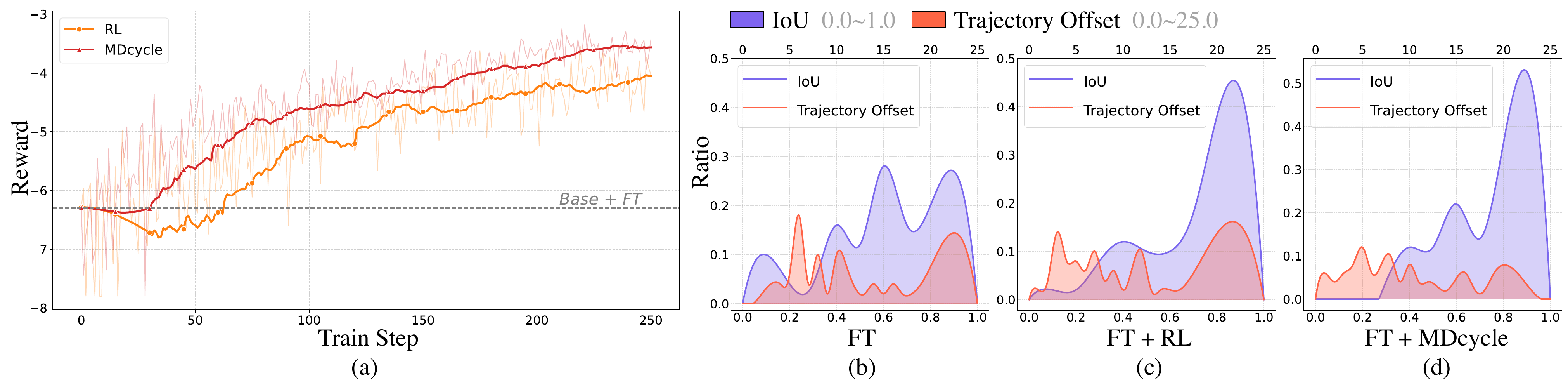}
    \vspace{-8mm}
    \caption{(a) \textbf{Reward Curve} of RL and \unipost. (b)(c)(d) \textbf{Metric Distribution} under different training strategies.
    }
    \label{fig:ablation_mdcycle}
    \vspace{-1mm}
\end{figure*}
%%%%%%%%%%%%%%%%%%%%%%%%%%%%%%%%%%%%%%%%%%%%

%%%%%%%%%%%%%%%%%%%%%%%%%%%%%%%%%%%%%%%%%%%%%%%%%
%##################################################################################################
% \renewcommand{\thesubtable}{{\alph{subtable}}}
\begin{table*}[t]
% \centering
% \hspace*{-0.05\linewidth}
\begin{minipage}[t]{1.1\linewidth}

% ######################  1  #######################
\subfloat[
]{
\begin{minipage}[t]{0.29\linewidth}
\tablestyle{4pt}{1.1}
\begin{tabular}{cccc}
\centering
SDE Window &Sample Step & IoU$\uparrow$ & TO$\downarrow$ \\
\hline
75\,\% -- 100\,\% &2 & \cellcolor{mygray}\textbf{0.64} & \cellcolor{mygray}\textbf{15.03} \\
50\,\% -- 100\,\% &2& 0.63 & 15.58 \\
25\,\% -- 100\,\% &2& 0.61 & 17.07 \\
\hline
0\,\% -- 100\,\%  &16& 0.55 & 27.24 \\
\end{tabular}
\label{subtab:sde_interval}
\end{minipage}
}
% \hspace{1em}
% ##################### 2 ######################
\subfloat[
]{
\begin{minipage}[t]{0.15\linewidth}
\tablestyle{4pt}{1.1}
\begin{tabular}{ccc}
\centering
$\sigma_{t}$ & IoU$\uparrow$ & TO$\downarrow$ \\
\hline
0.2      & 0.61 & 17.31 \\
0.6      & 0.64 & 15.83 \\
1.0      & \cellcolor{mygray}\textbf{0.64} & \cellcolor{mygray}\textbf{15.03} \\
1.4      & 0.62 & 16.91 \\
\end{tabular}
\label{subtab:eta}
\end{minipage}
}
% ################################################
% ###################### 3 ####################
\subfloat[
]{
\begin{minipage}[t]{0.18\linewidth}
\tablestyle{4pt}{1.1}
\begin{tabular}{ccc}
\centering
$\mathrm{Threshold}$ & IoU$\uparrow$ & TO$\downarrow$ \\
\hline
4       & 0.57 & 22.39 \\
8       & \cellcolor{mygray}\textbf{0.64} & \cellcolor{mygray}\textbf{15.03} \\
12      & 0.61 & 16.97 \\
\end{tabular}
\label{subtab:threshold}
\end{minipage}
}
% \hspace{1em}
% ################################################
% ##################### 4 ######################
\subfloat[
]{
\begin{minipage}[t]{0.25\linewidth}
\tablestyle{4pt}{1.1}
\begin{tabular}{ccc}
\centering
FT steps, MD steps & IoU$\uparrow$ & TO$\downarrow$ \\
\hline
(6000, 250)      & 0.49 & 37.23 \\
(16000, 250) & \cellcolor{mygray}\textbf{0.64} & \cellcolor{mygray}\textbf{15.03} \\
(30000, 250) & 0.64 & 15.08 \\
\end{tabular}
\label{subtab:training_schedule}
\end{minipage}
}
% ################################################
% ################################################
% ################################################
\end{minipage}
\vspace{-4mm}
\caption{\textbf{Hyperparameter Analysis.} All comparisons are conducted under the same data and GPU-hours budget. (a) SDE Interval. (b) Noise Intensity $\bm{\sigma}_{t}$. (c) Threshold in \cref{eq:md_cycle_loss}. (d) Training Steps. The best configuration is marked in \colorbox{mygray}{gray}.}
\label{tab:analysis}
\vspace{-4mm}
\end{table*}
%##################################################################################################

%%%%%%%%%%%%%%%%%%%%%%%%%%%%%%%%%%%%%%%%%%%%%%%%%

% \noindent \textbf{MDcycle.}
\label{sec:ablation_mdcycle}
\noindent \textbf{MDcycle and Other Training Strategies.} 
We conduct an ablation study to evaluate the core objective of \unipost, which is to stabilize reinforcement learning and improve the overall quality of generated samples. As shown in \cref{subtab:training_strategy} and \cref{fig:ablation_mdcycle}, \unipost achieves smoother convergence and higher final performance than other strategies. 
The reward curves in \cref{fig:ablation_mdcycle}(a) show that \unipost converges steadily and reaches higher rewards, while pure RL training remains unstable during the initial 50 steps and finally settles at a lower level. This instability arises because the model cannot produce reliable outputs within limited exploratory attempts, which restricts effective learning in early stages.
A similar pattern is observed in the metric distributions of sampled results, as shown in \cref{fig:ablation_mdcycle}(b,c,d). Compared with FT and FT+RL, \unipost increases the proportion of high-quality samples, such as those with TO in [0, 5] and IoU in [0.8, 1.0], and significantly reduces low-quality cases, such as those with TO in [15, 25] and IoU in [0.0, 0.4]. FT+RL already performs better than FT, and \unipost achieves the best results across all metrics. Together, these results demonstrate that \unipost provides greater training stability and stronger physical consistency than other optimization strategies.

\label{sec:ablation_collision}
\noindent \textbf{Collision Detection in Reward Function.}
We conduct an ablation study to evaluate the effect of collision-aware design in the reward model.
\cref{fig:ablation_collision} presents the qualitative results, where the FT model produces artifacts such as object disappearance and inaccurate trajectories, and RL improves motion smoothness but fails to handle collisions correctly. Incorporating collision-aware rewards eliminates these issues and yields physically consistent motions.
\cref{subtab:collision_weight} presents the results of varying the weighting factor $w$. The performance remains stable within a reasonable range, and we select the best value based on this study for all subsequent experiments.

\label{sec:ablation_sde_interval}
\noindent \textbf{SDE Interval.} We conduct a hyperparameter study on the hybrid SDE-ODE sampling strategy to evaluate its effect on RL training efficiency. Within each sampling window, two consecutive steps use an SDE sampler, and all remaining sampling steps use an ODE sampler. Results in \cref{subtab:sde_interval} show that SDE sampling performs best in high-noise regions, indicating that stochastic exploration at early noisy stages enhances semantic learning and overall stability.

\label{sec:ablation_sigma}
\noindent \textbf{Noise Intensity.}
$\sigma_{t}$ in~\cref{eq:sde} controls the noise intensity, which determines how extensively the model adjusts its parameters. In most cases, the value is set lower than the optimal $1.0$ reported in~\cref{subtab:eta} (\textit{e.g.}, $0.3$ in DanceGRPO), since excessive noise may degrade the model’s original performance and training stability.
In \method, we encourage active exploration of physical knowledge, making a higher noise ratio desirable. 
Stable training under this setting is achieved because the V2V process provides an unbiased initialization, ensuring reliable convergence.

\label{sec:ablation_threshold}
\noindent\textbf{Threshold.}
We analyze the \textit{Threshold} in \cref{eq:md_cycle_loss}, which regulates the balance between mimicry and discovery. As shown in \cref{subtab:threshold}, a small threshold causes over-mimicry and limits exploration, whereas a large one weakens guidance and destabilizes training. A proper threshold allows the model to adaptively transition from mimicry-driven stabilization to discovery-driven exploration, achieving a balanced and self-regulated learning process.

\label{sec:ablation_train_schedule}
\noindent \textbf{Training Steps.}
As shown in~\cref{subtab:training_schedule}, the benefit of full-parameter fine-tuning before the \unipost quickly saturates, with the optimal point around 16k steps. This indicates that the V2V process mainly performs data alignment at the visual level, ensuring a lower bound of visual quality, while the upper bound of physical knowledge learning still relies on the \unipost.

\section{Conclusion}
We present \method, a unified reinforcement learning framework that enhances the physical modeling ability of video generative models. Conventional pretrain–finetune paradigms focus on pixel reconstruction and perceptual quality while neglecting physical realism. To overcome this, we introduce a physics-grounded metric that measures motion fidelity and integrate abstract physical knowledge through \unipost. We also develop \bench, a benchmark for evaluating rigid-body motion quality. Extensive experiments demonstrate the effectiveness of our approach. Future works, limitations, and ethical considerations are provided in \textit{Appendix}.

{
    \small
    \bibliographystyle{ieeenat_fullname}
    \bibliography{main}
}

\setcounter{page}{1}
\maketitlesupplementary
\appendix

% ------------------------------------------------------------------------
\begin{algorithm*}[t]
\caption{MDcycle Training Algorithm}\label{algo:mdcycle}
\small
\begin{algorithmic}[1]
\Require Video generative policy model $\pi_\theta$; trajectory offset reward fucntion $R$; paired video-text datasets $\mathcal{D} = \{v^i,c^i\}_{i=1}^M$; coordinates of the first frame $\mathcal{P} = \{p^i\}_{i=1}^M$. SDE window timesteps $\mathcal{S} = \{s_i\}_{i=1}^K$; total sampling timestep $T$, threshold for \textit{Mimicry} branch $\mathrm{Threshold}$; 
\Ensure Optimized policy model $\pi_\theta$

\For{training iteration $=1$ \textbf{to} $N$}
    \State Sample batch $\mathcal{D}_b \sim \mathcal{D}$ and correspond coordinates $p \in \mathcal{P}$ 
    \State Update old policy: $\pi_{\theta_{\text{old}}} \gets \pi_\theta$
    
    \For{each context $v_{obs},c \in \mathcal{D}_b$ and correspond coordinates $p \in \mathcal{P}$:} 
        \State Generate $G$ samples: $\{\mathbf{x}_i\}_{i=1}^G \sim \pi_{\theta_{\text{old}}}(\cdot|v_{obs},c)$ with the same random initialization noise
        \For{each sample $\mathbf{x}$:} 
            \State Extract motion mask $\{m_i\}_{i=1}^T$ using \textit{SAM2} and $p$ 
            \State Extract trajectory ${\{p_i\}_{i=1}^T}$ using $Center$ operation
            \State Collision detect and get collision weight $\{w_t\}_{t=1}^T$
            \State Compute weighted trajectory offset (TO) $o_c$
            \State Get Reward $r \gets -o_c$ 
        \EndFor
        
        \State Gather group rewards $R \gets \{r_i\}_{i=1}^G$ and group trajectory $O_c \gets \{o_{c,i}\}_{i=1}^G$
        \State Compute group-average trajectory offset $\bar{O}_c \gets \frac{1}{G} \sum_{i=1}^G o_{c,i}$ 
        \State \dotfill
        \If{$\bar{O}_c > \mathrm{Threshold}$} \Comment{Mimicry Branch}
            \For{each sample $\mathbf{x}$:} 
                \State Calculate mimicry loss $l_M$ using \textbf{Flow Matching} algorithm
            \EndFor
            \State Gather group mimicry loss $L_M \gets \frac{1}{G}\sum_{i=1}^G l_M$ 
       
        \Else
            \State $L_M \gets0$
        \EndIf
        \State \dotfill
        \For{each sample $\textbf{x}$}: \Comment{Discovery Branch}
            \State Calculate advantage: 
            $a_i \gets \frac{r_i - \mu}{\sigma}$ 

            \For{SDE timestep window $t \in \mathcal{S} = \{s_i\}_{i=1}^K$} 
                \State Calculate discovery loss $l_D$ using \textbf{GRPO} algorithm
            \EndFor
            \State Gather group discovery loss $L_D \gets \frac{1}{G}\sum_{i=1}^G l_D$ 
        \EndFor

         \State Update policy via gradient descent:
            $\theta \gets \theta - \eta \nabla_\theta ( L_D + L_M)$
    \EndFor
\EndFor
\end{algorithmic}
\end{algorithm*}
% ------------------------------------------------------------------------

%%%%%%%%%%%%%%%%%%%%%%%%%%%%%%%%%%%%%%%%%%%%%
\section{Mimicry-Discovery Cycle Details}

\unipost builds on the reinforcement learning paradigm and is structured around two complementary components, the Mimicry Branch and the Discovery Branch. These components are integrated into a coherent framework that defines how the model alternates between data-driven imitation and physics-aware exploration.
A full description of this process, including step-by-step operations, is given in the pseudocode presented in \cref{algo:mdcycle}.

%%%%%%%%%%%%%%%%%%%%%%%%%%%%%%%%%%%%%%%%%%%%%
\section{\method Architecture \& Training Details}
\label{sec:network}

\noindent \textbf{\method Architecture.}
\method is built on the pre-trained video generation model Wan2.2 5B (TI2V). Leveraging its strong visual generative prior, we design a two-stage training pipeline. In Stage-1, we fully fine-tune the Wan2.2 5B TI2V model into a video-to-video model using a mixture of open-source and proprietary video collections. We keep the context length fixed at 5 frames. 
In Stage-2, we enhance the model with Physics-Aware generation capability by training \unipost on high-quality rigid-body motion data. During this stage, \unipost operates without full-parameter optimization to maintain stable learning.

\begin{figure}[t]
  \centering
    % \placeholder{8cm}{4cm}
    \includegraphics[width=1\linewidth]{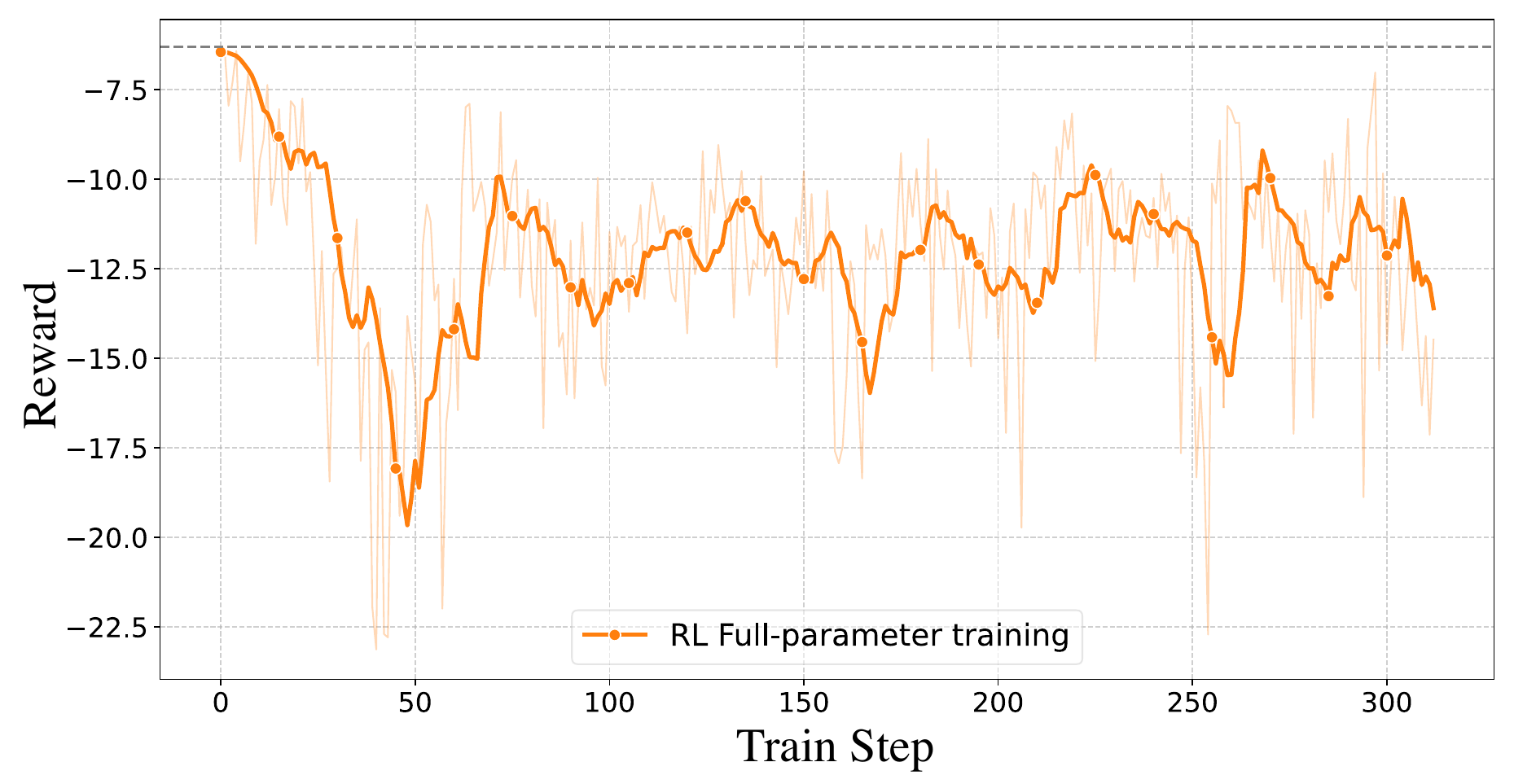}
    \vspace{-8mm}
    \caption{Reward curve of RL trained with full-parameter.}
    \label{fig:rl_fullparam}
    \vspace{-4mm}
\end{figure}

\noindent \textbf{Why We Avoid Full-Parameter Training in \unipost.}
Training in RL ranges from full-parameter updates to PEFT-style strategies (\textit{e.g.}, LoRA), and in our framework parameter efficient fine-tuning offers substantially better stability. As shown in \cref{fig:rl_fullparam}, full-parameter RL produces severe instability, and the reward curve collapses early and never recovers. This behavior is rooted in the high dimensionality of video generation, where reward signals are weak and ambiguous and the optimization landscape contains many misleading directions that push the model toward incorrect behaviors. Stable RL for video requires extremely large batch sizes, and even with our current configuration of 32 GPUs, one group per GPU, and 20 samples per group (effective batch size 640), the signal remains insufficient for reliable full-parameter training. Exploring larger-scale configurations is left for future work.

%%%%%%%%%%%%%%%%%%%%%%%%%%%%%%%%%%%%%%%%%%%%%%%%%
%%%%%%%%%%%%%%%%%%%%%%%%%%%%%%%%%%%%%%%%%%%%%%%%%

\begin{table}[t]
\centering
\small
\begin{tabular}{l|l}
\toprule[1.5pt]
\textbf{Parameter}              & \textbf{Value} \\ \midrule
Video Resolution                & 480*832            \\ 
Input Video Frames              & 49             \\ 
Input Video FPS                 & 30              \\ 
Optimizer                       & Adam; $\beta_1 = 0.9, \beta_2 = 0.95$ \\ 
Learning Rate                   & $1 \times 10^{-5}$ \\ 
Weight Decay                    & 0.0001 \\ 
Training Device                 & 32 $\times$ 80G H20 GPUs \\
Train steps /\textit{stage1}     & 16000 \\
Train steps /\textit{stage2}     & 250 \\
Batchsize /\textit{stage1}       & 32 \\
Batchsize /\textit{stage2}       & $32(gpu)* 20(sample) = 640$ \\ \midrule 
Group per GPU                   & 1  \\
Samples per Group               & 20 \\
Noise intensity $\sigma_{t}$    & 1.0 \\
$(w,w^{adj},w^{col})$           & 1,2,3 \\
SDE window size                 & 2  \\
SDE window interval             & 75\% - 100\% \\
Threshold                       & 8 \\
Sampling Steps                  & 16 \\
Same initial noise          & True \\
CFG                             & False \\

\bottomrule[1.5pt]
\end{tabular}
\vspace{-2mm}
\caption{Training Configuration}
\vspace{-6mm}
\label{tab:training_config}
\end{table}

\noindent \textbf{Hyperparameters and Environment.}
All experiments are conducted on 4 nodes, each equipped with 8 H20 GPUs. We train the model for 16,000 steps in Stage 1 and 250 steps in Stage 2. It is worth noting that we do not use classifier-free guidance (CFG) during the sampling phase of \unipost. The main reasons are that CFG roughly doubles the computational cost and can introduce training instability. In our V2V setting, the model benefits from rich contextual signals in the input video, which allows us to discard CFG without significantly harming visual quality. However, we do not recommend removing CFG in T2V tasks, where such strong contextual information is absent and turning off CFG typically leads to a substantial degradation in sample quality. Within each group, we use the same noise initialization for all samples. Under the constraint of a relatively small batch size, this shared-noise strategy further contributes to stabilizing RL training. All of our training configurations are listed in~\cref{tab:training_config}.

%%%%%%%%%%%%%%%%%%%%%%%%%%%%%%%%%%%%%%%%%%%%%
\section{Benchmark Details}
\subsection{Video Samples}
\begin{figure*}[!t]
  \centering
    % \placeholder{8cm}{4cm}
    \includegraphics[width=1\linewidth]{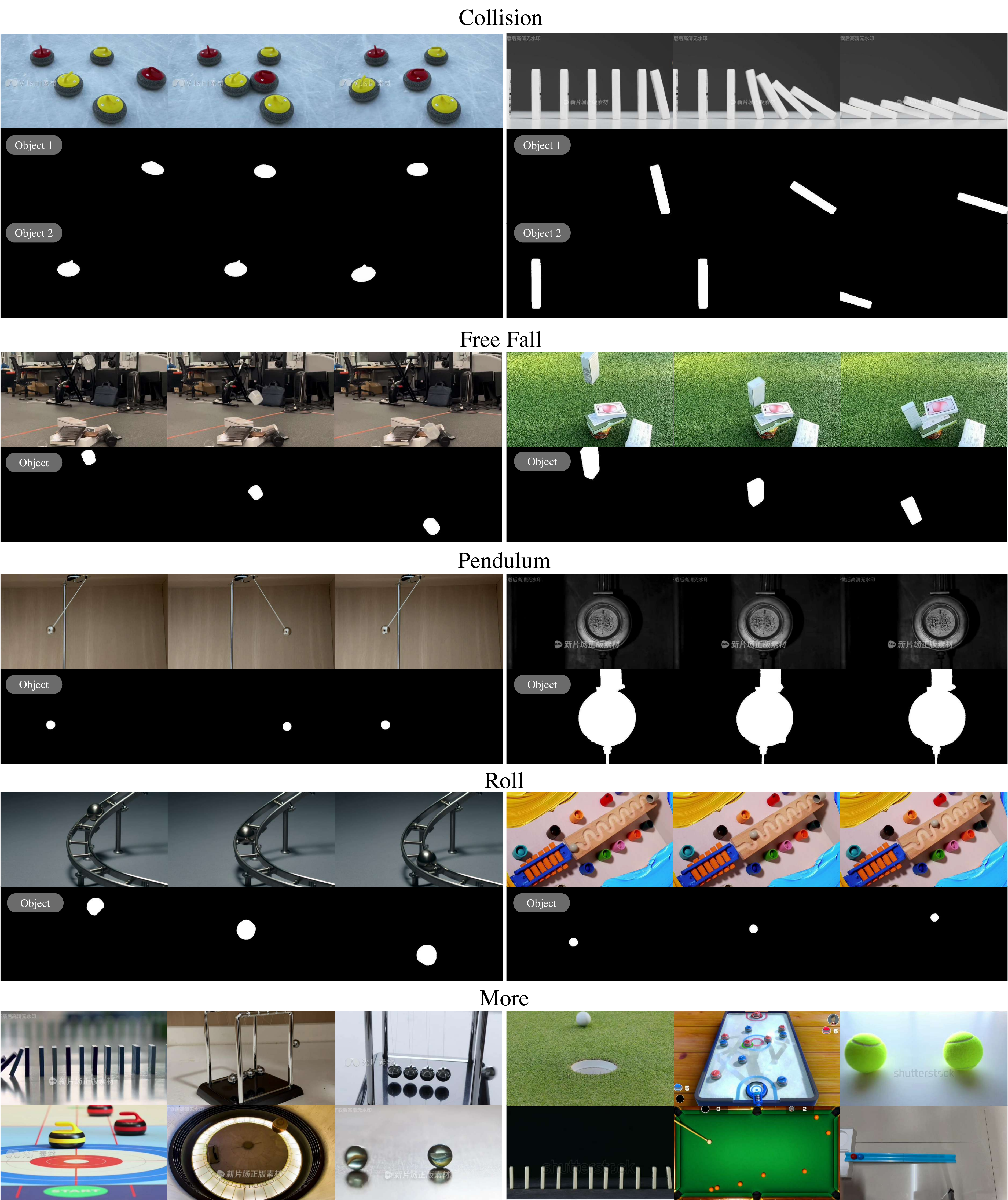}
    \caption{Videos in the \bench.}
    \label{fig:supp_benchmark_data}
    \vspace{-1mm}
\end{figure*}

In this section, we present a detailed description of the dataset construction process. Representative samples are provided in~\cref{fig:supp_benchmark_data}. Our \bench focuses on four types of rigid-body motion: collision, free fall, rolling, and pendulum motion. The data are collected from several sources, including existing datasets (\textit{e.g.}, PISA~\cite{li2025pisa}), the Internet, video game recordings, and real-world experiments captured using our own devices. For every sample, the text prompt is fixed to ``\textit{The video shows rigid body motion}'', ensuring that the model relies primarily on the context frames rather than textual variation. We manually annotate object coordinates in the first frame. In collision scenes, annotations include both the active object and the passive object, while in other scenarios we annotate only the primary moving object. Based on the first-frame coordinates, we use SAM2~\cite{ravi2024sam2} to extract motion masks. In collision scenes, although two objects are involved, we segment them in two separate passes because generating both masks at once often produces noticeably less accurate results.

%%%%%%%%%%%%%%%%%%%%%%%%%%%%%%%%%%%%%%%%%%%%
%%%%%%%%%%%%%%%%%%%%%%%%%%%%%%%%%%%%%%%%%%%%

\subsection{Evaluation Metrics}
In \bench, we employ two metrics to evaluate the physical realism of rigid-body motion: interaction over Unions (IoU) and our newly proposed trajectory offset (TO). The IoU metric measures the overlap between the predicted and ground-truth interaction regions, while the trajectory offset evaluates the discrepancy between the generated motion trajectory and the ground-truth trajectory. By combining these two metrics, we obtain a more comprehensive assessment of the physical plausibility of rigid-body motion. In the following, we describe in detail how these two metrics are computed.

%%%%%%%%%%%%%%%%%%%%%%%%%%%%%%%%%%%%%%%%%%%%%%%%%
\begin{figure}[h]
  \centering
    % \placeholder{8cm}{4cm}
    \includegraphics[width=1\linewidth]{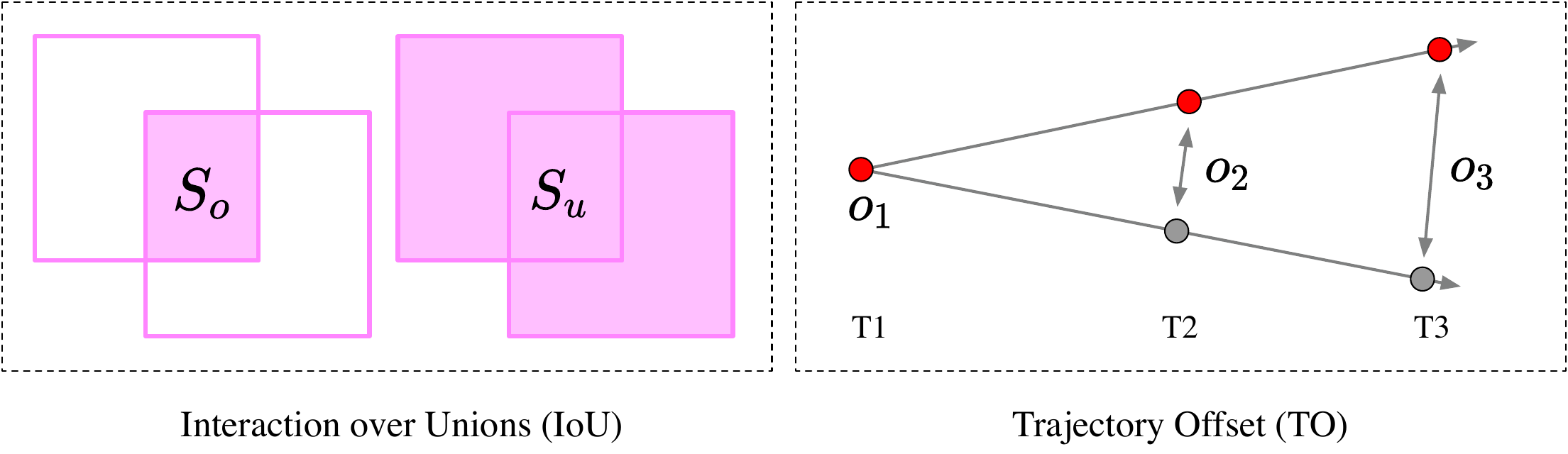}
    \caption{Evaluation Metric.}
    \label{fig:evaluation_metric}
\end{figure}
%%%%%%%%%%%%%%%%%%%%%%%%%%%%%%%%%%%%%%%%%%%%%%%%%

\begin{itemize}
    \item \textbf{interaction over Unions (IoU):} As illustrated in~\cref{fig:evaluation_metric}, Intersection over Union (IoU) is a metric used to measure the overlap between two regions. It is defined as the area of their overlap $S_o$ divided by the area of their union $S_u$. Formally, the IoU is given by:
    $\mathrm{IoU} = \frac{S_o}{S_u}.$

    \item \textbf{Trajectory Offset (TO):} As illustrated in~\cref{fig:evaluation_metric}, For each timestamp $T_i$, we extract the ground-truth object coordinates $p_i^{gt}$ and the generated object coordinates $p_i^{sample}$, and compute the distance between the two $o_t =\left\| p_{t}^{\text{gt}} - p_{t}^{\text{sample}} \right\|_2$. We then average these per-frame distances over all frames to obtain the final trajectory offset $O = \frac{1}{T} \sum_{t=T_{obs}}^{T} \left\| p_{t}^{\text{gt}} - p_{t}^{\text{sample}} \right\|_2$.

\end{itemize}

%%%%%%%%%%%%%%%%%%%%%%%%%%%%%%%%%%%%%%%%%%%%%
\section{Detection Method}
In this section, we describe how we perform collision detection. Our goal is to identify the time indices of collisions based on the motion mask extracted by SAM2~\cite{ravi2024sam2}. The core idea is inspired by Newton’s second law, $F=ma$: assuming the object’s mass remains constant, the moment when an external force is applied corresponds to a change in acceleration. Therefore, we detect collisions by locating the time points at which the acceleration increases sharply. In practice, we detect abrupt changes in acceleration using functions from the \textit{SciPy} Python library. The corresponding pseudocode is provided in~\cref{algo:supp_collision}.

\begin{figure}[t]
\begin{algorithm}[H]
% \small
\caption{\small Collision Detection Algorithm}
\label{algo:supp_collision}
\definecolor{codeblue}{rgb}{0.1,0.6,0.1}
% \definecolor{codekw}{rgb}{0.85, 0.18, 0.50}
\definecolor{codekw}{rgb}{0.85, 0.18, 0.50}
\lstset{
  backgroundcolor=\color{white},
  basicstyle=\fontsize{7.5pt}{7.5pt}\ttfamily\selectfont,
  columns=fullflexible,
  breaklines=true,
  captionpos=b,
  commentstyle=\fontsize{7.5pt}{7.5pt}\color{codeblue},
  keywordstyle=\fontsize{7.5pt}{7.5pt}\color{codekw},
  escapechar={|}, 
}
\begin{lstlisting}[language=python]
from scipy.signal import find_peaks
import numpy as np

# Require: 
# Video: v
# Coordinate: p

# Motion Mask
masks = SAM2(v,p)     

# Coordinates list
coords = Center(masks)

# Velocity,Acceleration
vels = np.diff(coords, axis=0)
accs = np.diff(vels, axis=0)

# Identify acceleration peaks
peaks, properties = find_peaks(accs,prominence,distance=distance)

# Output Collision indexs
collision_frames = peaks+2 
# We cannot obtain the acceleration for the first two frames, so we add an offset of 2 here.
return collision_frames

\end{lstlisting}
\label{alg1}
\end{algorithm}
% \vspace{-30pt}
\vspace{-2mm}
\end{figure}

\begin{figure}[t]
  \centering
    % \placeholder{8cm}{4cm}
    \includegraphics[width=1\linewidth]{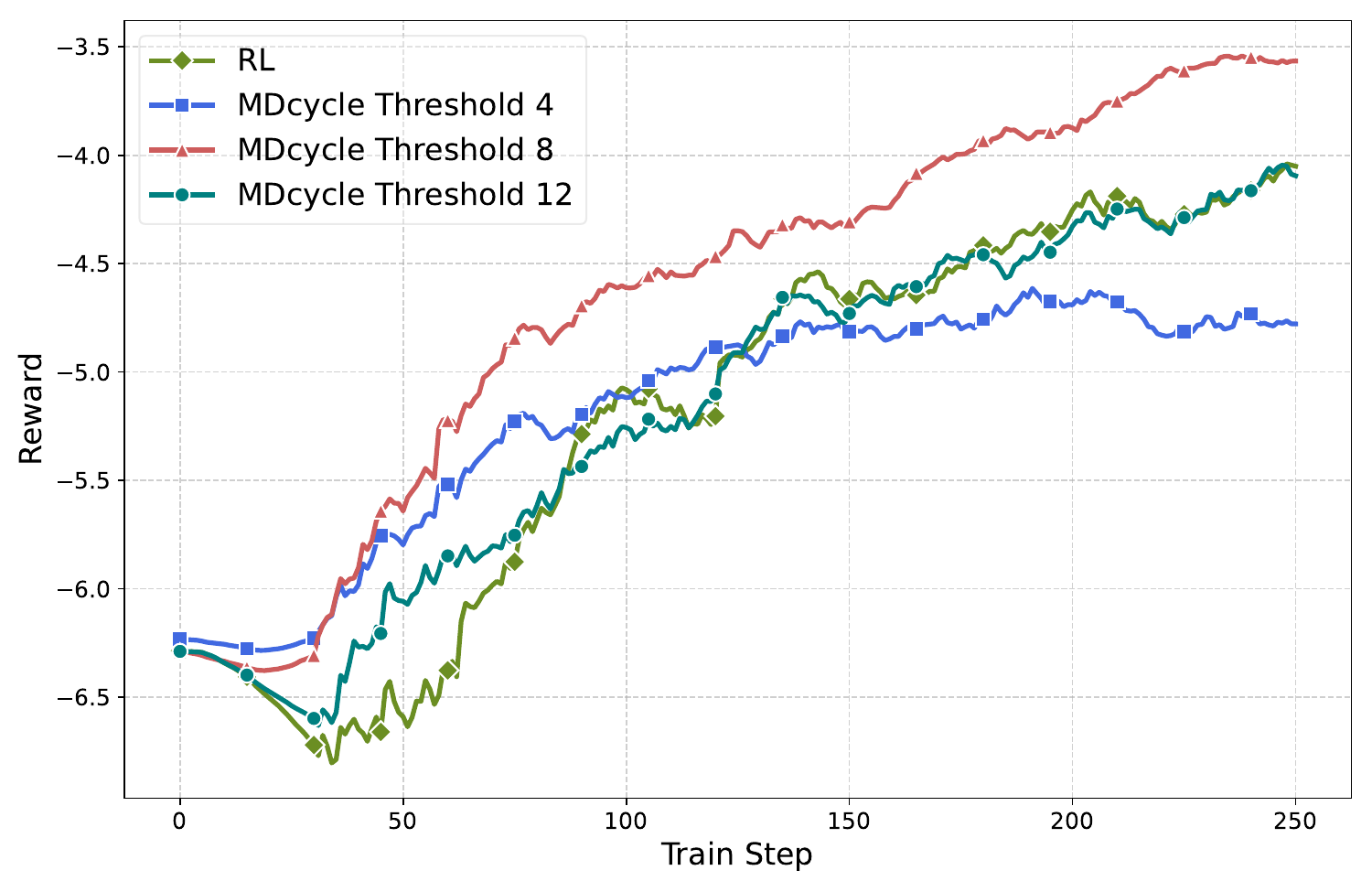}
    \vspace{-8mm}
    \caption{Reward curve of different Threshold settings.}
    \label{fig:supp_comparison_loss_curve}
\end{figure}

%%%%%%%%%%%%%%%%%%%%%%%%%%%%%%%%%%%%%%%%%%%%
\begin{figure*}[t]
  \centering
    % \placeholder{8cm}{4cm}
    \includegraphics[width=0.9\linewidth]{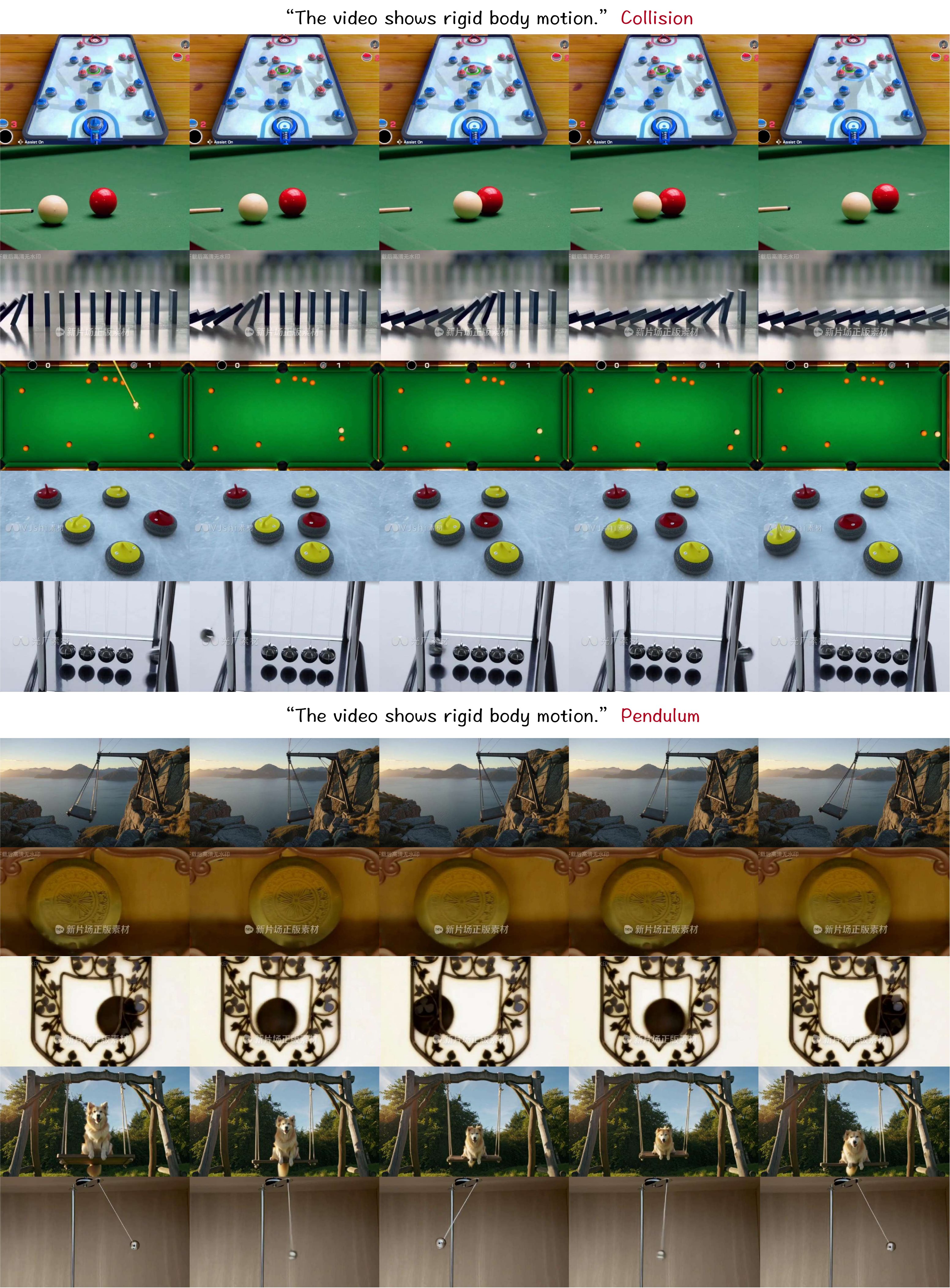}
    \caption{More results generated by \method
    }
    \label{fig:supp_case_collision}
    \vspace{-1mm}
\end{figure*}
%%%%%%%%%%%%%%%%%%%%%%%%%%%%%%%%%%%%%%%%%%%%

%%%%%%%%%%%%%%%%%%%%%%%%%%%%%%%%%%%%%%%%%%%%
\begin{figure*}[t]
  \centering
    % \placeholder{8cm}{4cm}
    \includegraphics[width=0.9\linewidth]{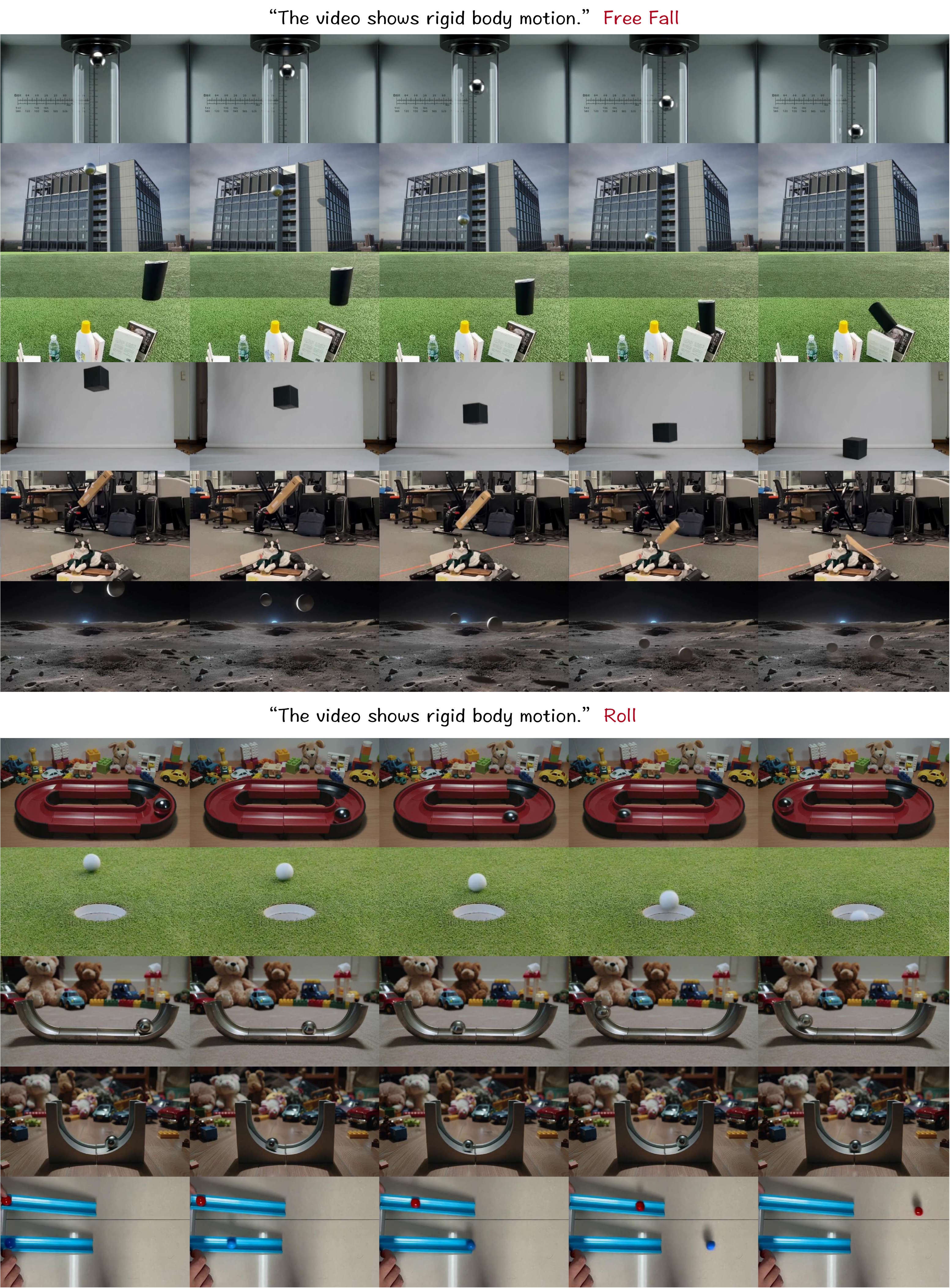}
    \caption{More results generated by \method
    }
    \label{fig:supp_case_freefall}
    \vspace{-1mm}
\end{figure*}
%%%%%%%%%%%%%%%%%%%%%%%%%%%%%%%%%%%%%%%%%%%%

%%%%%%%%%%%%%%%%%%%%%%%%%%%%%%%%%%%%%%%%%%%%%
\section{More Generated Results.}\cref{fig:supp_case_collision} and~\cref{fig:supp_case_freefall} present additional results generated by our method. All context frames are out-of-distribution: they are either taken from the evaluation set or generated by Kling2.5 Turbo. In collision scenarios, we observe that the model has learned to produce complex motions involving multiple objects and multiple collision events. In pendulum scenarios, although the training data only contain simple pendulum setups (e.g., a clock pendulum or a single swinging ball), the model can generalize this behavior to similar scenes such as a swinging playground swing. In free fall scenarios, the model not only produces realistic falling motion but also reasonably captures the subsequent impact with the ground. In rolling scenarios, the model is able to perform repeated acceleration and deceleration along the track, consistent with the track geometry and energy conservation.

%%%%%%%%%%%%%%%%%%%%%%%%%%%%%%%%%%%%%%%%%%%%%
\section{More analysis on Threshold.}
In this section, we further analyze the Threshold. The Threshold controls the degree to which the Mimicry branch participates in training. Although involving the Mimicry branch generally improves training stability, more participation is not always better. As shown in~\cref{fig:supp_comparison_loss_curve}, when the Threshold is very small (e.g., threshold = 4 in the reward curve), the Mimicry branch helps stabilize training in the early stage, but ultimately leads to premature convergence to a lower reward. We believe this happens because, in the later stages of the \unipost, the model needs more freedom for RL-based exploration, and excessive pixel-level supervision from the Mimicry branch hampers exploration and constrains the model’s performance ceiling.

Conversely, when the threshold is very large, the convergence behavior becomes almost identical to pure RL, as illustrated in the figure. When the threshold is set to a moderate, appropriate value, the model not only converges more stably but also achieves a higher final reward. Although this hybrid training strategy is effective, the choice of hyperparameters is largely empirical and must be tuned for different tasks. In future work, we plan to design an automatic adjustment mechanism that can balance training stability and exploration capability.

%%%%%%%%%%%%%%%%%%%%%%%%%%%%%%%%%%%%%%%%%%%%%
\section{Discussion}
\label{sec:discussion}

\subsection{Limitation}
\begin{figure}[t]
  \centering
    % \placeholder{8cm}{4cm}
    \includegraphics[width=1\linewidth]{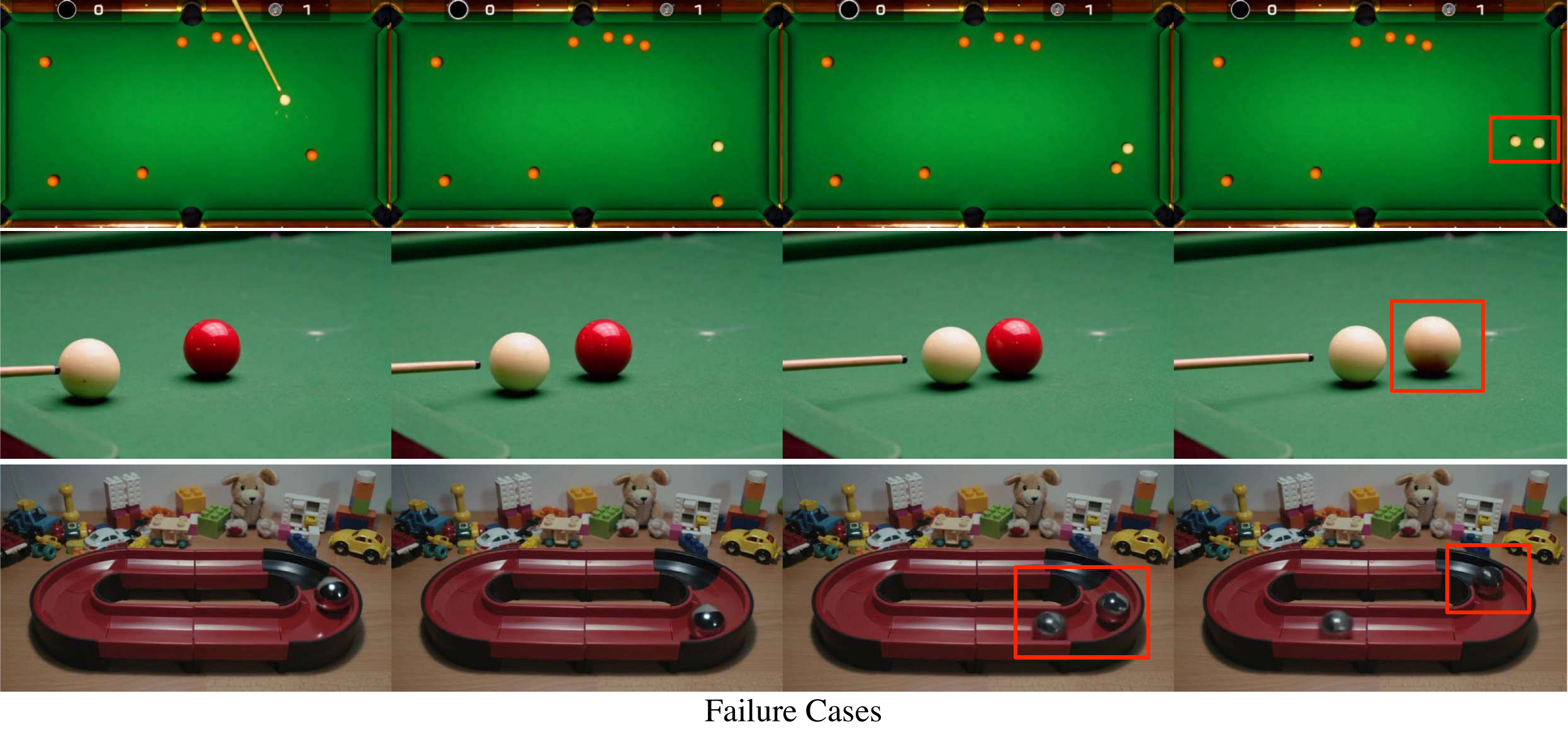}
    \vspace{-6mm}
    \caption{Failure cases generated by \method
    }
    \label{fig:supp_faulure}
    \vspace{-1mm}
\end{figure}

Although our model can generate videos with highly realistic motion trajectories, it still makes mistakes in aspects that are weakly correlated with the primary motion. For example, as shown in~\cref{fig:supp_faulure}, the objects in row 1 and row 2 change color after collision, and in row 3, an extra ball appears in the frame when the original ball turns. We attribute these issues to inherent limitations of the model, as these undesirable cases are not supervised by our reward. Our reward is solely related to the motion trajectory of the moving object and is not directly concerned with its color, the presence or behavior of other objects in the scene, or object shape. As a result, the model is not penalized when such bad cases occur. This suggests that we need a more comprehensive, multi-scale evaluation framework for assessing the physical realism of object motion, which is extremely challenging.

\subsection{Ethical Considerations}
\label{sec:ethical}
Our work is motivated by improving the physical realism of generative models~\cite{banerjee2024physicsinformedcomputervisionreview,liu2025generativephysicalaivision,meng2025groundingcreativityphysicsbrief,hu2025simulatingrealworldunified}, enabling constructive applications from scientific simulation and robotics to enhanced creative pipelines in film and games. However, we recognize that any advance that makes synthetic media more plausible also heightens the potential for misuse. A system capable of generating physically consistent videos could be exploited to create highly convincing disinformation, such as fabricated accidents or false evidence, that circumvents detection methods relying on physical inconsistencies. Furthermore, while our work focuses on rigid objects, the underlying techniques are transferable and could be misapplied to create defamatory or privacy-invasive content by placing real individuals into synthetically generated, compromising physical situations. As creators of this technology, we acknowledge our responsibility to mitigate these risks. Our reinforcement learning framework, while optimizing for physical accuracy, does not inherently encode ethical or safety constraints, making it agnostic to the content it generates. Therefore, we are committed to a responsible release, which includes clearly labeling our models for research purposes, documenting potential misuse scenarios, and encouraging the adoption of content provenance and watermarking technologies. 

%%%%%%%%%%%%%%%%%%%%%%%%%%%%%%%%%%%%%%%%%%%%
%%%%%%%%%%%%%%%%%%%%%%%%%%%%%%%%%%%%%%%%%%%%

\end{document}